\pdfoutput=1

\documentclass[11pt]{article}
\usepackage[final]{acl}

\usepackage{times}
\usepackage{latexsym}

\usepackage[T1]{fontenc}
\usepackage[utf8]{inputenc}
\usepackage{microtype}
\usepackage{inconsolata}
\usepackage{graphicx}
\usepackage{todonotes}

\usepackage{makecell}
\usepackage{enumitem}

\usepackage{arydshln}
\usepackage{adjustbox}
\usepackage{multirow}
\usepackage{multicol}
\usepackage{booktabs}
\usepackage{tabularx}

\newcommand{\llamap}[0]{Llama2$_{C}$}
\newcommand{\mistralp}[0]{Mistral$_{I}$}
\newcommand{\llama}[0]{\llamap\ }
\newcommand{\mistral}[0]{\mistralp\ }
\title{Should We Fine-Tune or RAG? \\ Evaluating Different Techniques to Adapt LLMs for Dialogue}

\makeatletter
\def\@fnsymbol#1{
    \ensuremath{
        \ifcase#1\or \dagger\or \ddagger\or \mathsection\or \mathparagraph\or \|\or **\or \dagger\dagger \or \ddagger\ddagger \else\@ctrerr\fi
    }
}
\makeatother

\author{
    Simone Alghisi\textsuperscript{\ $\dagger$\ }, Massimo Rizzoli\thanks{Equal contribution.}, Gabriel Roccabruna, \\ 
    \textbf{Seyed Mahed Mousavi, Giuseppe Riccardi} \\
     Signals and Interactive Systems Lab, University of Trento, Italy \\
    \texttt{ \{s.alghisi, massimo.rizzoli, giuseppe.riccardi\}@unitn.it}
}

\begin{document}
\maketitle

\begin{abstract}
We study the limitations of Large Language Models (LLMs) for the task of response generation in human-machine dialogue.
Several techniques have been proposed in the literature for different dialogue types (e.g., \textit{Open-Domain}). 
However, the evaluations of these techniques have been limited in terms of base LLMs, dialogue types and evaluation metrics.
In this work, we extensively analyze different LLM adaptation techniques when applied to different dialogue types. 
We have selected two base LLMs, \llama and \mistralp, and four dialogue types \textit{Open-Domain}, \textit{Knowledge-Grounded}, \textit{Task-Oriented}, and \textit{Question Answering}. 
We evaluate the performance of in-context learning and fine-tuning techniques across datasets selected for each dialogue type. 
We assess the impact of incorporating external knowledge to ground the generation in both scenarios of Retrieval-Augmented Generation (RAG) and gold knowledge. 
We adopt consistent evaluation and explainability criteria for automatic metrics and human evaluation protocols. 
Our analysis shows that there is no universal best-technique for adapting large language models as the efficacy of each technique depends on both the base LLM and the specific type of dialogue. 
Last but not least, the assessment of the best adaptation technique should include human evaluation to avoid false expectations and outcomes derived from automatic metrics.

\end{abstract}

%\todo[inline, color=blue!20]{text}
%\todo[color=green!40,inline,caption={}]{}

\section{Introduction}
In recent years, Large Language Models (LLMs) have been employed for the task of response generation in human-machine dialogues~\citep{hosseini-asl-2020-simpletod, izacard-grave-2021-leveraging, komeili-etal-2022-internet}.
Such models have been applied to several dialogue types, including Open-Domain Dialogues (i.e. informal conversations about trivial matters), Knowledge-Grounded Dialogues (i.e. conversations with a system that provides factual responses), Task-Oriented Dialogues (i.e. conversations where the system helps a user to achieve a specific goal), and Question Answering (i.e. question-answer exchanges given context).

However, recent studies have shown the shortcomings of LLMs as dialogue model surrogates as they are prone to generate toxic, biased, and irrelevant responses~\citep{zhang-etal-2020-dialogpt, mousavi-etal-2022-evaluation, mousavi-etal-2023-response, lin-chen-2023-llm}.
To adapt LLMs to dialogue types, different techniques have been employed such as in-context learning~\citep{NEURIPS2020_1457c0d6, chen-etal-2023-exploring-context, meade-etal-2023-using, hudecek-dusek-2023-large} and fine-tuning~\citep{wang-2022-todnlg, komeili-etal-2022-internet, huang-etal-2023-learning}.
Furthermore, strategies such as grounding~\citep{gopalakrishnan-2019-topical, zhao-etal-2023-others} and Retrieval-Augmented Generation (RAG)~\citep{lewis-2020-retrieval, borgeaud2022improving} have been proposed to improve the generation quality.

Currently, the performance of the aforementioned techniques in adapting LLMs across different dialogue types is understudied. Previous studies have evaluated these techniques in a specific dialogue type only \citep{raposo-etal-2023-prompting, zhang-etal-2023-survey-efficient}. Such studies are based on different base models and are assessed via incomparable evaluation methodologies.

In this work, we conduct an extensive study on the efficacy of different techniques to adapt LLMs for multiple dialogue types.
We select Llama-2 Chat (\llamap)~\citep{touvron2023llama} and Mistral Instruct (\mistralp)~\citep{jiang2023mistral} as base LLMs, and experiment with in-context learning and fine-tuning in the context of four dialogue types: a) Open-Domain Dialogues (ODDs), b) Knowledge-Grounded Dialogues (KGDs), c) Task-Oriented Dialogues (TODs), d) Question Answering (QA).
Besides, we assess the impact of incorporating external knowledge by considering retrieved knowledge and gold knowledge.
In the retrieved knowledge scenario, we use RAG to add the knowledge to the model's input.
We assess the performance of each technique using the same automatic metrics and comparable human evaluation.
We further compute the contribution of each segment of the input vector by using integrated gradients as an explainability attribution method.
We evaluate the models using an open human evaluation protocol~\cite{mousavi-etal-2022-evaluation} designed for dialogue contextualization, appropriateness, correctness, and validity. In summary, the main contributions of this paper are:

\begin{itemize}
     \item Adaptation of \llama and \mistral using fine-tuning and in-context learning\footnote{The code is available at \url{https://github.com/sislab-unitn/Fine-Tune-or-Rag}} in four different dialogue types and corresponding corpora;
     \item Assessment of the impact of grounding the response generation on external knowledge, both in cases of retrieved knowledge and gold knowledge;
     \item Extensive study on the efficacy of each technique using automatic evaluations and human evaluation, including explainability and categorization analysis of natural language generation errors.

\end{itemize}

\section{Literature Review}

\textbf{Open-Domain Dialogue (ODD)} 
In earlier studies, sequence-to-sequence models have been trained for response generation in open-domain dialogues~\cite{li-etal-2017-dailydialog}. However, such models suffered from generating generic or inappropriate responses ~\citep{zhang-etal-2020-dialogpt}. 
To improve the generation quality, studies grounded the generation on external knowledge, such as persona statements~\citep{wolf-2019-transfertransfo, kasahara-etal-2022-building, xu-etal-2022-long}, the personal graph of user interactions~\citep{mousavi-etal-2023-response}, and retrieved documents~\citep{huang-etal-2023-learning}. 
While the previous works developed data-driven models using training/fine-tuning, recent studies have explored the potential of in-context learning with LLMs~\citep{qian-etal-2023-harnessing}.

\textbf{Knowledge-Grounded Dialogue (KGD)} 
Sources such as Wikipedia have been used as unstructured knowledge to ground the generated responses~\citep{dinan-2019-wizard, gopalakrishnan-2019-topical,komeili-etal-2022-internet} to generate consistent and factual answers.
To improve the generation quality, previous works have studied the impact of knowledge selection~\citep{qin-etal-2023-well,sun-etal-2023-generative}, different knowledge representations~\citep{mousavi-etal-2023-response,yang-etal-2023-graph}, additional knowledge elements (e.g. dialogue acts, topics)~\citep{hedayatnia-etal-2020-policy}, training without knowledge supervision~\citep{han-etal-2023-efficient}, and in-context learning~\citep{chen-etal-2023-exploring-context}.

\textbf{Task-Oriented Dialogue (TOD)}
LLMs have been fine-tuned for TOD modeling for joint dialogue state tracking and response generation~\citep{hosseini-asl-etal-2020-simpletod, kulhanek-etal-2021-augpt, wang-2022-todnlg, ding-etal-2024-kmctod}, and robustness to spoken interactions~\cite{thulke-2024-tod-dstc9-10, mousavi2024llms}.
Recent studies focus on augmenting the TOD modeling with unstructured knowledge access ~\citep{feng-etal-2020-doc2dial, kim-etal-2020-beyond, kim-2021-dstc10}.
In this regard, \citet{he-etal-2023-dstc9winner} have proposed a pipeline for retrieval and grounded response generation.
\citet{raposo-etal-2023-prompting} compared in-context-learning and fine-tuning, but considered retrieved replies from previous dialogues as knowledge.

\textbf{Question Answering (QA)}.
In the most general setting, relevant documents need to be retrieved to provide an answer~\citep{lee-etal-2019-latent, qu-2020-open}.
Some studies have proposed to select the documents with the highest similarity with the question computed between their BERT encodings~\citep{lee-etal-2019-latent, karpukhin-etal-2020-dense}.
With this retrieval strategy, some studies have fine-tuned LLMs to condition the generation on the retrieved documents through grounding~\citep{lewis-2020-retrieval, izacard-grave-2021-leveraging} or cross-attention~\citep{borgeaud2022improving}.
Other works generated the answers using in-context learning with zero-shot~\cite{levine2022huge, cho-etal-2023-improving}.
A survey compared existing generation-only, retrieval-only, and RAG models~\citep{zhang-etal-2023-survey-efficient} but with different base models, hindering the comparison of the techniques.

\section{Experiments}
We study and compare in-context learning and fine-tuning as techniques to adapt LLMs for human-machine dialogues.
We select Llama-2 Chat (\llamap)~\citep{touvron2023llama} and Mistral Instruct (\mistralp)~\citep{jiang2023mistral} as base LLMs, and experiment in the context of four dialogue types: Open-Domain Dialogue (ODD), Knowledge-Grounded Dialogue (KGD), Task-Oriented Dialogue (TOD), and Question Answering (QA). 
For each technique and dialogue type, we assess the impact of grounding the generation on documents in the scenarios of retrieved knowledge (RAG) and gold knowledge.

\subsection{Datasets}
In our experiment, we have selected a dataset for each of the four dialogue types (see §\ref{sec:app_datasets} for selection). 
The statistics of these datasets are summarized in Table~\ref{tab:datasets}.

\textbf{Open-Domain Dialogue (ODD)}
We select DailyDialog~\cite{li-etal-2017-dailydialog}, a widely-used dataset of human-human dialogues crawled from various websites used by English learners to practice. 
The final dataset contains 13k written dialogues with an average of 8 turns per dialogue.

\textbf{Knowledge-Grounded Dialogue (KGD)}
We experiment on Wizard of Wikipedia ~\citep{dinan-2019-wizard}, a dataset of dialogues between two participants with the roles of apprentice and wizard.
At each turn, the wizard can access a set of documents (passages from Wikipedia) and use it to incorporate factual knowledge in their reply.
The dataset contains 20k dialogues about one of 1359 distinct topics and provides an unseen set of documents for testing.

\textbf{Task-Oriented Dialogue (TOD)} 
We select the dataset proposed for the first track of the ninth Dialogue System Technology Challenge~\citep{kim-etal-2020-beyond}, an augmented version of MultiWOZ 2.1~\citep{eric-etal-2020-multiwoz}.
The dataset spans over 7 domains and contains 9k multi-domain dialogues. 
The dialogues include turns where the system needs to access an unstructured knowledge base of 2900 documents (FAQs) to provide a correct response.

\textbf{Question Answering (QA)}
We select NarrativeQA~\citep{kocisky-etal-2018-narrativeqa}, a dataset of 47k questions with free-form answers based on 1.5k books and movie scripts. 
The question-answer pairs are formulated based on summaries of the books and movies.

\begin{table}
\small
    \centering
    \begin{tabular}{llrrr}
        \toprule
        \textbf{Type} & \textbf{Dataset} & \textbf{\#Dials} & \textbf{\makecell{Avg. \\ \#Turns}} & \textbf{\makecell{\#Ext. \\ Know.}} \\
        \midrule
        \textit{ODD} & DailyDialog & 13k & 8 & --- \\
        \textit{KGD} & WoW & 20k & 9 & \textsuperscript{\textdagger}61 \\
        \textit{TOD} & DSTC9 Track 1 & 9k & 19 & 2900 \\
        \textit{QA} & NarrativeQA & \textsuperscript{*}47k & 2 & 1572 \\
        \bottomrule
        \end{tabular}
        \caption{
            Selected datasets for each dialogue type: Open-Domain Dialogue (ODD), Knowledge-Grounded Dialogue (KGD), Task-Oriented Dialogue (TOD), and Question Answering (QA).
            \#Ext. know. indicates the number of documents in the unstructured knowledge base.
            \textsuperscript{\textdagger} In KGD the content of the knowledge base differs at each turn with an average of $61 \pm 22$ documents.
            \textsuperscript{*} Question-answer exchanges.
        }
    \label{tab:datasets}
\end{table}

\subsection{Techniques}
We evaluate in-context learning and fine-tuning as techniques to adapt LLMs for response generation in the selected dialogue types. 
In-context learning is a technique that uses instructions and examples to condition the generation.
Instead, fine-tuning further trains the model (completely or partially) on the task of interest using a smaller-scale dataset than the pre-training phase.
In a dialogue setting, fine-tuning should {\em teach} LLMs to behave as dialogue models and account for each state of the conversation between speakers.

As a baseline, for both techniques, we consider the context (i.e. the question for QA, the history for ODD, KGD, and TOD) as the input and use the default prompt structure of the models to separate user and system turns.
Additionally, for TOD we append the dialogue state (a summary of user requirements), following previous work on this dialogue type~\citep{wang-2022-todnlg, ding-etal-2024-kmctod}.
For KGD, we prepend the topic to the start of the dialogue.

\begin{table*}[t!]
\small
    \centering
    \begin{tabularx}{0.91\linewidth}{lllcccc}
        \toprule
        \multirow{2}{*}{\textbf{Model}} & \multirow{2}{*}{\textbf{Technique}} & \multirow{2}{*}{\textbf{\makecell{External\\Knowledge}}} & \multicolumn{4}{c}{\textbf{Perplexity}} \\
        \cmidrule(rrrr){4-7}
        & & & \texttt{ODD} & \texttt{KGD} & \texttt{TOD} & \texttt{QA} \\
        \midrule
        \multirow{6}{*}{\textbf{\llama}} & \multirow{3}{*}{\textit{In-Context Learning}} & \texttt{No Know.} & 64.13 & 35.17 & 25.15 & 1442.26 \\
        & & \texttt{Retrieved Know.} & & 33.10 & 24.72 & 625.08 \\
        & & \texttt{Gold Know.} & & 24.40 & 23.81 & 298.16 \\
        \cmidrule{2-7}
        & \multirow{3}{*}{\textit{Fine-Tuning}} & \texttt{No Know.} & \textbf{5.67 \tiny{$\pm$ 0.01}} & 7.63 \tiny{$\pm$ 0.01} & \textbf{3.06 \tiny{$\pm$ 0.01}} & 12.03 \tiny{$\pm$ 0.06} \\
        & & \texttt{Retrieved Know.} & & 6.95 \tiny{$\pm$ 0.01} & 3.97 \tiny{$\pm$ 0.01} & 5.47 \tiny{$\pm$ 0.02} \\
        & & \texttt{Gold Know.} & & \textbf{4.38 \tiny{$\pm$ 0.01}} & 3.12 \tiny{$\pm$ 0.01} & \textbf{4.98 \tiny{$\pm$ 0.01}} \\
        \midrule
        \multirow{6}{*}{\textbf{\mistral}} & \multirow{3}{*}{\textit{In-Context Learning}} & \texttt{No Know.} & 14.19 & 15.31 & 9.82 & 91.42 \\
        & & \texttt{Retrieved Know.} & & 14.75 & 9.76 & 42.58 \\
        & & \texttt{Gold Know.} & & 9.81 & 9.37 & 16.74 \\
        \cmidrule{2-7}
        & \multirow{3}{*}{\textit{Fine-Tuning}} & \texttt{No Know.} & \textbf{6.41 \tiny{$\pm$ 0.01}} & 8.67 \tiny{$\pm$ 0.01} & \textbf{3.56 \tiny{$\pm$ 0.01}} & 14.11 \tiny{$\pm$ 0.01} \\
        & & \texttt{Retrieved Know.} & & 7.78 \tiny{$\pm$ 0.01} & 3.61 \tiny{$\pm$ 0.01} & 5.97 \tiny{$\pm$ 0.01} \\
        & & \texttt{Gold Know.} & & \textbf{5.17 \tiny{$\pm$ 0.01}} & 3.58 \tiny{$\pm$ 0.01} & \textbf{4.88 \tiny{$\pm$ 0.01}} \\
        \bottomrule
    \end{tabularx}
    \caption{\textbf{Automatic Evaluation} Perplexity of Fine-Tuning and In-Context Learning with \texttt{Retrieved} (top-3) and \texttt{Gold} (ground-truth) knowledge, on \llama and \mistralp, in different dialogue types: Open-Domain Dialogues (ODDs), Knowledge Grounded Dialogues (KGDs), Task-Oriented Dialogues (TODs), and Question Answering (QA). Results for fine-tuned models report mean and standard deviation over three runs.}
    \label{tab:ppl}
\end{table*}

\subsection{Knowledge}
Incorporating external knowledge for the task of response generation has been shown to improve the factual accuracy~\citep{he-etal-2023-dstc9winner} and contextualization~\citep{mousavi-etal-2023-response} of responses.

For each of the selected types but for ODD, we consider their corresponding unstructured knowledge base.
Regarding KGD, we consider passages from Wikipedia, while for TOD we consider FAQs related to services and places (e.g. restaurants, hotels, taxi booking).
For QA we consider all the summaries of the books and movies.

For both in-context learning and fine-tuning, we study the impact of knowledge on the generated responses, in two scenarios:
\begin{itemize}
    \item \textbf{Retrieved knowledge}: we retrieve k documents from the unstructured knowledge base;
    \item \textbf{Gold knowledge}: we use the ground truth document.
\end{itemize}

For the retrieved knowledge scenario, we use the Retrieval Augmented Generation (RAG) strategy.
We use an off-the-shelf retriever\footnote{\url{https://github.com/langchain-ai/langchain}} (model details in §\ref{sec:app_impl_and_perf}) to retrieve documents from the unstructured knowledge base.
First, we encode all the documents considering their content together with their topic (KGD), place or service name (TOD), or title (QA)~\citep{karpukhin-etal-2020-dense}.
Then, at each turn, we retrieve the k most similar documents based on L2 distance with the encoded context.
Finally, we feed the retrieved documents to the base models together with the context to generate a response.

In the gold knowledge scenario, we directly feed the model with the ground truth documents.
This serves as an upper bound for RAG.
Additionally, this strategy allows us to study the ability of the techniques to incorporate knowledge in the responses.

\subsection{Models}
We select the widely-used 7B version of \llama and \mistral as base models.
For in-context learning, we experiment with three instructions for each dialogue type and select the best based on the development set performance.
For fine-tuning, we use LoRA, a parameter-efficient technique that has shown comparable performance to fine-tuning all parameters~\citep{hu-2021-lora}. 
Further details about the parameters are reported in §\ref{sec:app_impl_and_perf}.

\section{Evaluation}
We conduct a comparative study on the impact of in-context learning and fine-tuning to adapt LLMs for dialogues.
We select \llama and \mistral as base LLMs and experiment in four dialogue types: ODDs, KGDs, TODs, and QA.
For each dialogue type, we study the impact of external knowledge, both retrieved and gold.
Further details about the implementation and the resources used are available in the appendix (§\ref{sec:app_impl_and_perf}).

\begin{table*}[!t]
\small
    \centering
    \begin{tabular}{lclcccc}
        \toprule
        \multirow{3}{*}{\textbf{Model}} & \multirow{3}{*}{\textbf{\makecell{Dialogue\\Type}}}  & \multirow{3}{*}{\textbf{Technique}} & \multicolumn{4}{c}{\textbf{\% of Tokens w. Significant Contribution in Each Segment}} \\
        \cmidrule{4-7}
        & & & \texttt{Instruction} & \texttt{\makecell{Topic/Dialogue\\ State}} & \texttt{\makecell{Dialogue\\History}} & \texttt{Knowledge} \\
        \midrule
        
        \multirow{4}{*}{\makecell{\textbf{\llama}}} & \multirow{2}{*}{\textit{KGD}} & In-Context Learning & 21.85 & 28.60 & 15.97 & \textbf{33.58} \\
        \cdashline{3-7}\noalign{\smallskip}
        & & Fine-Tuning & & 39.43 & 13.80 & \textbf{46.77} \\
        \cmidrule{2-7}
        & \multirow{2}{*}{\textit{TOD}} & In-Context Learning & 25.98 & 19.54 & 16.46 & \textbf{38.02} \\
        \cdashline{3-7}\noalign{\smallskip}
        & & Fine-Tuning & & 27.19 & 8.04 & \textbf{64.77} \\
        \cmidrule{1-7}
        \multirow{4}{*}{\makecell{\textbf{\mistral}}} & \multirow{2}{*}{\textit{KGD}} & In-Context Learning & & \textbf{69.01} & 14.89 & 16.10 \\
        \cdashline{3-7}\noalign{\smallskip}
        & & Fine-Tuning & & \textbf{65.55} & 11.00 & 23.45 \\
        \cmidrule{2-7}
        & \multirow{2}{*}{\textit{TOD}} & In-Context Learning & \textbf{69.05} & 10.19 & 11.24 & 9.52 
        \\ \cdashline{3-7}\noalign{\smallskip}
        & & Fine-Tuning & & 14.55 & 29.06 & \textbf{56.39} \\

        \bottomrule
        \end{tabular}
        \caption{
            \textbf{Explanability Study} Percentage of tokens with significant contribution to the generation in different segments of the input vector for each model in Knowledge-Grounded Dialogues (KGDs), and Task-Oriented Dialogues (TODs). 
            All rows sum to 100. 
            For KGD, the second column reports the contribution of the \texttt{Topic}, while for TOD it reports the contribution of the \texttt{Dialogue State}.
            The \texttt{Instruction} segment is only present for In-Context Learning. 
        }
    \label{tab:expl_kgd_tod}
\end{table*}

\subsection{Automatic Evaluation}
\label{sec:auto-eval}
Currently available automatic metrics used for the task of response generation are not interpretable and correlate poorly with human judgments~\citep{liu-etal-2016-evaluate, sai2022surveyevalnlg, mousavi-etal-2022-evaluation}.
Therefore, we focus on perplexity as it is derived from the objective function used to fine-tune the models, and present other metrics in §\ref{sec:add_autom_eval}.

Table \ref{tab:ppl} reports the perplexity of \llama and \mistral on the test set of each dialogue type.
In all dialogue types, fine-tuned models have obtained better performance compared to in-context learning.
When considering the impact of external knowledge, models fine-tuned on TODs show that knowledge slightly increases perplexity.
The high perplexity obtained by in-context learning models on QA can be explained by two reasons: first, besides the knowledge, only the question is used as context; second, while the ground truths are particularly short (4.26 tokens on average), these models generate long responses, making them unlikely to include the correct answer in the first few tokens.
This does not happen for fine-tuned models since they are trained to generate shorter responses.
Nevertheless, the best results have been obtained with gold knowledge.
We report automatic evaluation results including retriever accuracy, overlap between knowledge and response tokens, and other automatic metrics in §\ref{sec:add_autom_eval}.

\subsubsection{Explainability Study}
To understand the contribution of each segment of the input vector (i.e. instruction, context, knowledge, topic, and dialogue state), we compute integrated gradients~\citep{sarti-etal-2023-inseq}\footnote{We use Inseq to compute integrated gradients.} of input elements and select the most contributing input tokens (top-25\%).
Table \ref{tab:expl_kgd_tod} reports the percentage of most contributing tokens that fall in each segment (normalized by the length of the segment). 
In general, in both KGD and TOD, the dialogue history is the least contributing segment, which might indicate that only a part of the history is significant for response generation.
On the other hand, in KGD the topic has a higher score than the dialogue history, suggesting its importance for response generation for this dialogue type.
Interestingly, \mistral gives considerably more importance to the topic than \llamap, decreasing the importance of the knowledge segment.
For the TOD type, the most contributing segment is often the knowledge, reaching over 50\% with fine-tuning.
This suggests that knowledge is more relevant for TOD and that relevance changes with respect to the dialogue type.

\begin{table*}[t!]
\small
    \centering
    \begin{tabularx}{\linewidth}{lXXcccccccc}
        \toprule
        \multirow{2}{*}{\textbf{Model}} & \multirow{2}{*}{\textbf{Technique}} & \multirow{2}{*}{\textbf{\makecell{External\\Knowledge}}} & \multicolumn{4}{c}{\textbf{Contextualization}} & \multicolumn{3}{c}{\textbf{Appropriateness}} & \textbf{Validity} \\
        \cmidrule(rrrr){4-7} \cmidrule(rrrr){8-10} \cmidrule(rrrr){11-11}
        & & & \texttt{ODD} & \texttt{KGD} & \texttt{TOD} & \texttt{QA} & \texttt{ODD} & \texttt{KGD} & \texttt{TOD} & \texttt{QA} \\
        \midrule
        \multirow{6}{*}{\textbf{\llama}} & \multirow{3}{*}{\textit{In-Context Learning}} & \texttt{No Know.} & \textbf{85} & 70 & 70 & 50 & \textbf{80} & 70 & 60 & 10 \\
        & & \texttt{Retrieved Know.} & & 75 & 65 & 70 &  & 75 & 45 & 35 \\
        & & \texttt{Gold Know.} & & \textbf{90} & 40 & \textbf{90} & & \textbf{85} & 45 & \textbf{80} \\
        \cmidrule{2-11}
        & \multirow{3}{*}{\textit{Fine-Tuning}} & \texttt{No Know.} & 45 & 60 & 70 & 15 & 50 & 65 & 60 & 15 \\
        & & \texttt{Retrieved Know.} & & 65 & \textbf{90} & 45 & & 80 & 80 & 45  \\
        & & \texttt{Gold Know.} & & 80 & 85 & 85 & & 65 & \textbf{85} & 75 \\
        \midrule
        \multirow{6}{*}{\textbf{\mistral}} & \multirow{3}{*}{\textit{In-Context Learning}} & \texttt{No Know.} & \textbf{90} & 80 & 70 & 20 & \textbf{85} & \textbf{85} & 65 & 20 \\
        & & \texttt{Retrieved Know.} & & 75 & 65 & 40 & & 65 & 60 & 25 \\
        & & \texttt{Gold Know.} & & 90 & 55 & \textbf{75} & & 70 & 55 & \textbf{80} \\
        \cmidrule{2-11}
        & \multirow{3}{*}{\textit{Fine-Tuning}} & \texttt{No Know.} & 55 & 90 & \textbf{85} & 25 & 55 & 80 & 80 & 20 \\
        & & \texttt{Retrieved Know.} & & \textbf{95} & \textbf{85} & 30 & & \textbf{85} & \textbf{90} & 40 \\
        & & \texttt{Gold Know.} & & 80 & 75 & 70 & & 65 & 70 & 70 \\
        \midrule
        \textbf{Ground-Truth} & & & 95 & 80 & 95 & 90 & 100 & 85 & 95 & 90 \\
        \bottomrule
        \end{tabularx}
        \caption{
            \textbf{Human Evaluation}
            Percentage of Contextualized, Appropriate (ODD, KGD, TOD), and Valid (QA) responses for In-Context Learning and Fine-Tuning with \texttt{Retrieved} (top-3) and \texttt{Gold} (ground-truth) knowledge, on \llama and \mistralp, in different dialogue types: Open-Domain Dialogues (ODDs), Knowledge Grounded Dialogues (KGDs), Task-Oriented Dialogues (TODs), and Question Answering (QA). 
        }
    \label{tab:human_eval}
\end{table*}

\subsection{Human Evaluation}
Considering the uninterpretability of automatic evaluations, we conducted a human evaluation of the generated responses to gain more insight into the models' performance.
\citet{mousavi-etal-2022-evaluation} proposed four dimensions to evaluate response generation based on the most common errors and qualities.
We evaluate the responses using their protocol and three of their dimensions:
\begin{itemize} [noitemsep]
    \item \textbf{Contextualization}: the response includes explicit or implicit references to the dialogue history (ODD, KGD, TOD) or the gold knowledge (QA);
    \item \textbf{Appropriateness}: the response is coherent and makes sense as a continuation of the dialogue;
    \item \textbf{Correctness}: the response is grammatically and syntactically correct.
\end{itemize}
According to these dimensions, we evaluate the responses for all techniques, models, and knowledge scenarios, in all dialogue types.
The only exception is QA, where we do not evaluate "Appropriateness" since the dimension considers coherence with respect to a dialogue history but QA only has question-answer exchanges.
Instead, we extend the protocol\footnote{The extended protocol is available at \url{https://github.com/sislab-unitn/Human-Evaluation-Protocol/tree/v1.1}} by proposing a new dimension for QA: 
\begin{itemize}[noitemsep]
    \item \textbf{Validity}: the response includes adequate information to answer the question.
\end{itemize}
For TOD we do not include a dimension to evaluate whether the response is in line with user requirements, as this can be measured automatically (via dialogue state tracking metrics e.g., Joint Goal Accuracy).
The dimensions can either have a positive or negative answer value, as well as "I don't know" to avoid forcing erroneous judgments on any of the two sides.
For "Contextualization" and "Appropriateness", we also ask the annotators to motivate the negative judgments with the explanations proposed in the original protocol.
We present the explanations and related results in §\ref{sec:expl_hum_judg}.

We recruited 75 annotators on the Prolific platform\footnote{\url{https://www.prolific.com/}}, and we assigned 5 dialogues to each annotator.
After performing quality control, we approved 65 annotators with a compensation of 9.00£/hour (marked as good on the Prolific platform). 
Due to the large number of responses, each annotator evaluated a different set of model responses for a given dialogue.
For the purpose of quality control, for each dialogue type, two dialogues were overlapping among five annotators, while the remaining dialogues were annotated by one crowd-worker with an overlap only on the ground truth.
The inter-annotator agreement measured with Fleiss' $\kappa$ \citep{fleiss1971measuring} was 0.65 (substantial agreement).

As results of the human evaluation (Table \ref{tab:human_eval}), we report the percentage of positively judged responses (Contextualized, Appropriate, Valid) for \llama and \mistral when considering different adaptation techniques (Fine-Tuning and In-Context Learning) and knowledge (No Knowledge, Retrieved Knowledge, and Gold Knowledge) across different dialogue types.
As for ODDs, we report no results for the Retrieved and Gold Knowledge scenarios since no knowledge was used for this dialogue type.
Additional results on "Correctness" are reported in §\ref{sec:app_hum_eval}.

%% ODD

\textbf{Open-Domain Dialogue (ODD)} Models fine-tuned for ODD tend to generate considerably less contextualized responses than models adapted using in-context learning.
In particular, fine-tuning \llama reduces contextualization by 40\%, while for \mistral by 35\%.
Similarly, fine-tuning reduces their appropriateness by 30\% compared to their in-context learning version.
This contrasts with automatic evaluation (Table \ref{tab:ppl}), where in-context learning obtained a higher perplexity (i.e. worse results) compared to fine-tuning.

%% KGD

\textbf{Knowledge-Grounded Dialogue (KGD)}
Concerning KGD, the results are model-dependent.
When considering \llamap, in-context learning provides, regardless of the knowledge, 10\% more contextualized responses compared to fine-tuning.
On the other hand, fine-tuning \mistral on Retrieved Knowledge leads to the highest contextualization (95\%).
However, using Gold instead of Retrieved Knowledge reduces the contextualization of the fine-tuned model by 15\%.
Furthermore, when considering the best models, \llama and \mistral have a higher contextualization than the ground truth (10 to 15\%), suggesting that models copy more from the dialogue history.
Similarly to contextualization, adapting \llama with in-context learning and Gold Knowledge provides the highest percentage of appropriate responses (85\%).
Instead, fine-tuning (on Retrieved Knowledge) or adapting \mistral with in-context learning (using No Knowledge) provides comparable appropriateness (85\%).
While according to automatic evaluation (Table \ref{tab:ppl}) fine-tuning is always the best technique, human evaluation results show comparable appropriateness and contextualization for in-context learning and fine-tuning.

\begin{figure}[t]
    \centering
    \includegraphics[width=\columnwidth]{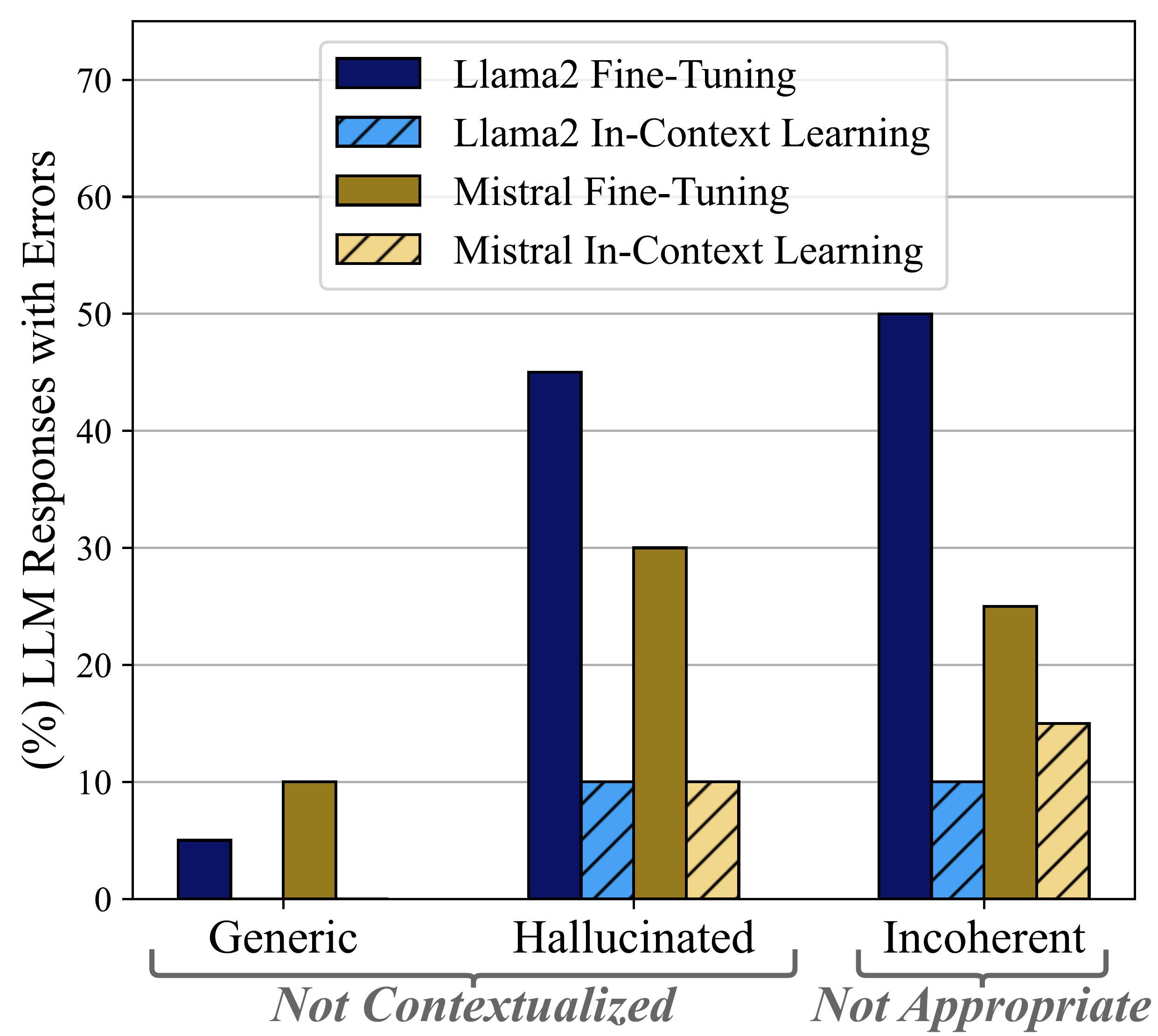}
    \caption{
        Percentage of LLM responses (y-axis) for each error type (\textit{Not Contextualized} and \textit{Not Appropriate}) and their explanation (Generic, Hallucinated, and Incoherent) (x-axis), for \llama and \mistralp, adapted with In-Context Learning and Fine-Tuning in Open-Domain Dialogues (ODDs).
    }
    \label{fig:odd_hum_err}
\end{figure}

\begin{figure}[t]
    \centering
    \includegraphics[width=\columnwidth]{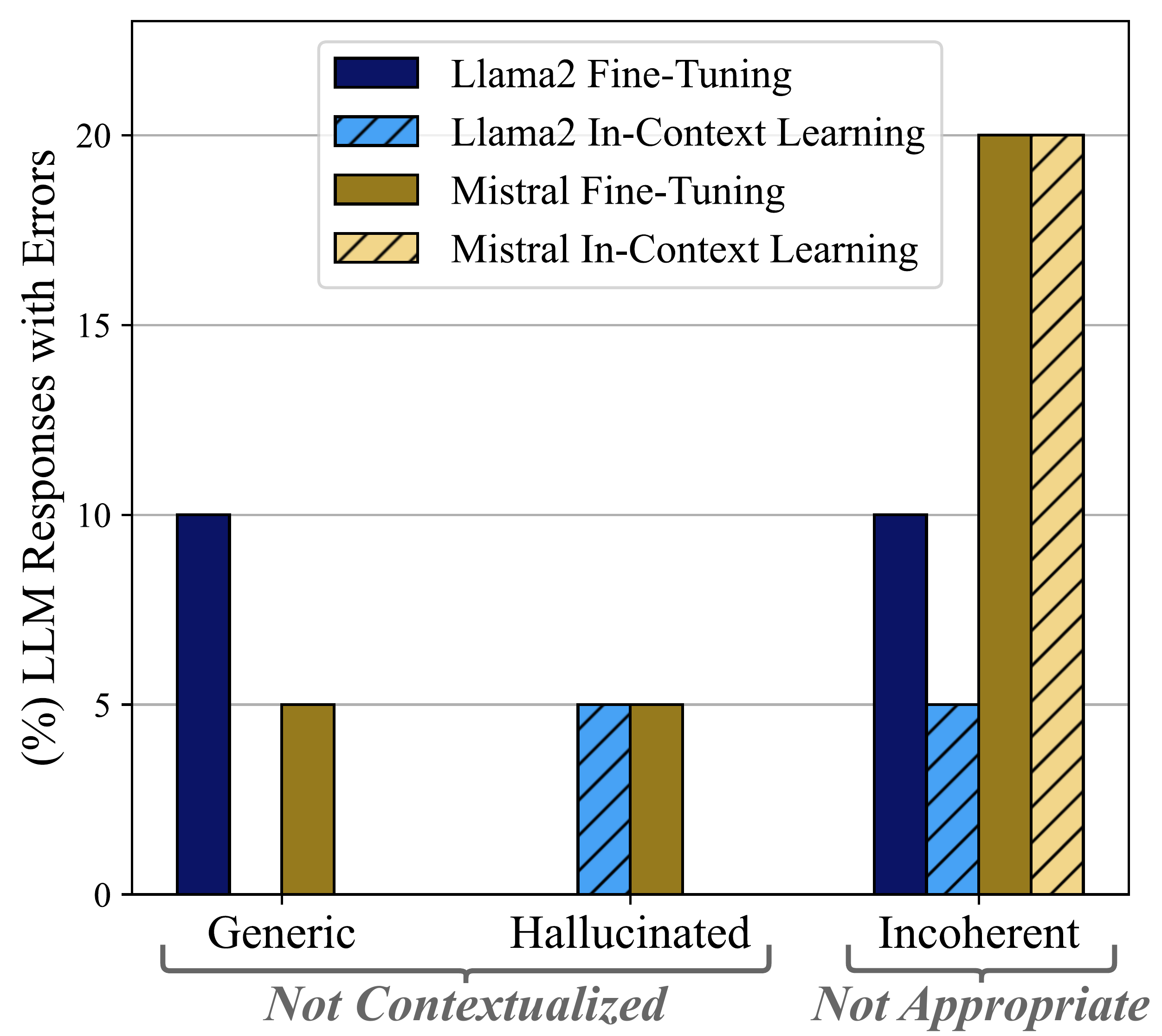}
    \caption{
        Percentage of LLM responses (y-axis) for each error type (\textit{Not Contextualized} and \textit{Not Appropriate}) and their explanation (Generic, Hallucinated, and Incoherent) (x-axis), for \llama and \mistralp, adapted with In-Context Learning and Fine-Tuning in Knowledge-Grounded Dialogues (KGDs).
    }
    \label{fig:kgd_hum_err}
\end{figure}

% TOD

\textbf{Task-Oriented Dialogue (TOD)} 
When adapting \llama and \mistral to TOD, the results clearly show that fine-tuning is preferable over in-context learning.
In particular, if we consider the best model for each technique, when fine-tuned \llama generates 20\% more contextualized responses, while \mistral generates 15\% more.
Although fine-tuned models benefit from external knowledge, Retrieved and Gold Knowledge visibly reduce contextualization of in-context learning models (at most by 30\% for \llama and 15\% for \mistralp).
Similar behavior can be observed for in-context learning in terms of appropriateness, where Gold Knowledge reduces \llama results by 15\% and \mistral by 10\%.
This is in line with the explainability study (Table \ref{tab:expl_kgd_tod}), where models adapted with in-context learning have a lower contribution from the knowledge segment than their fine-tuned version. 
In general, if we consider the best models for each technique, fine-tuned models generate 25\% more appropriate responses.

% QA

\textbf{Question Answering (QA)}
In QA, results show improved contextualization and validity when including knowledge, with the best results obtained with gold knowledge.
When considering the best model for each technique, in-context learning increases the percentage of contextualized responses by 5\%.
These results greatly differ from Table \ref{tab:ppl} and show how unreliable automatic evaluation can be.
Although models fine-tuned on No or Retrieved Knowledge obtain comparable or higher validity than in-context learning, adding Gold Knowledge to adapt \llama and \mistral with in-context learning increases their validity respectively by 5\% and 10\%.
Finally, even with Gold Knowledge, no model reaches the validity of the ground truth (90\%).

% summary

These findings indicate that the best technique depends on the dialogue type and the base LLM.
Regarding the techniques, in-context learning leads to more contextualized and appropriate responses in ODDs, while fine-tuning improves contextualization and appropriateness in TODs.
Regarding the base LLMs, in KGDs adapting \llama with in-context learning leads to the best results, while \mistral benefits the most from fine-tuning.
Furthermore, in QA the quality of knowledge impacts contextualization and validity the most, while adaptation techniques have a minor effect.

\begin{figure}[t]
    \centering
    \includegraphics[width=\columnwidth]{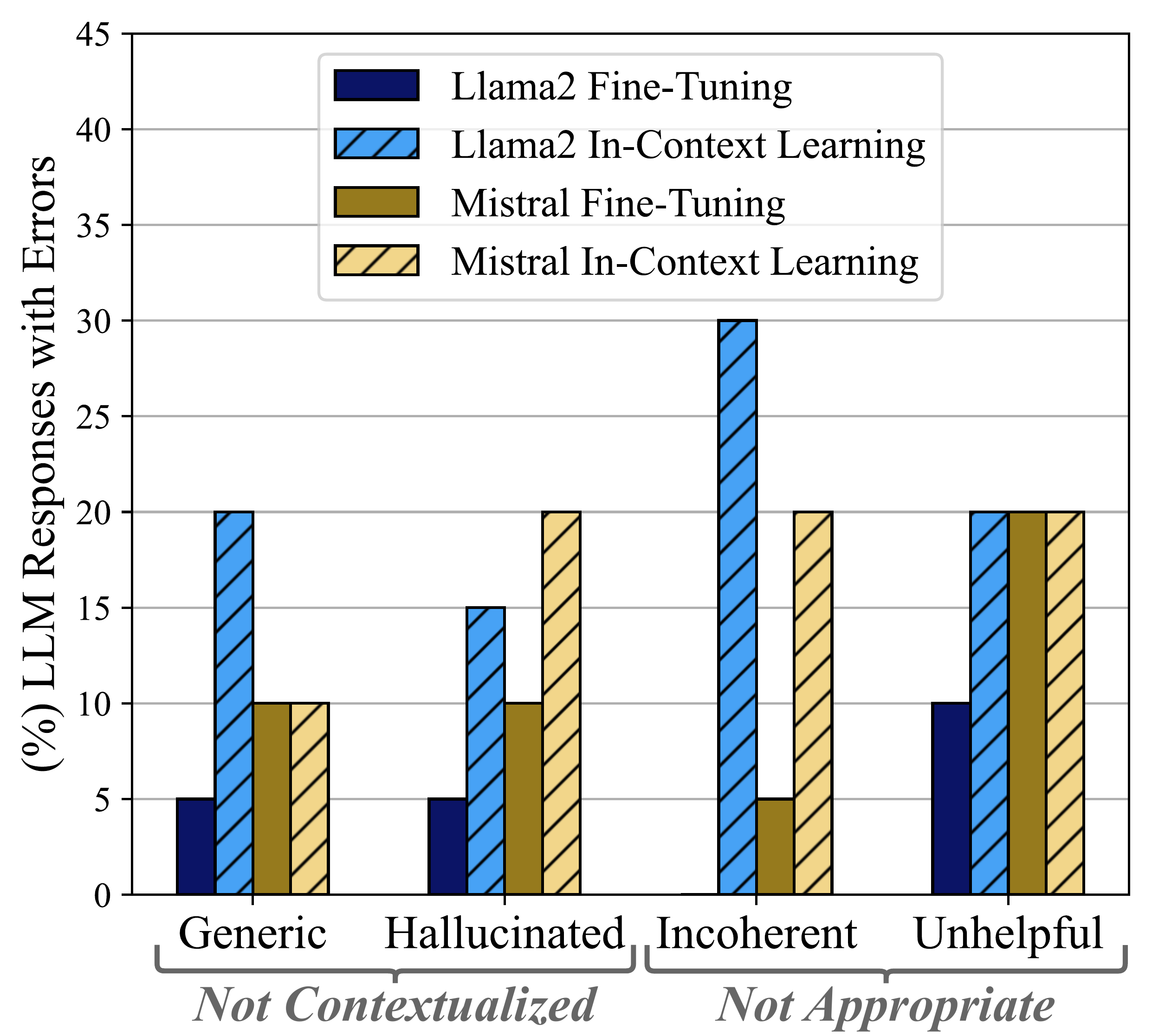}
    \caption{
        Percentage of LLM responses (y-axis) for each error type (\textit{Not Contextualized} and \textit{Not Appropriate}) and their explanation (Generic, Hallucinated, Incoherent, and Unhelpful) (x-axis), for \llama and \mistralp, adapted with In-Context Learning and Fine-Tuning in Task-Oriented Dialogues (TODs).
    }
    \label{fig:tod_hum_err}
\end{figure}

\begin{figure}[t]
    \centering
    \includegraphics[width=\columnwidth]{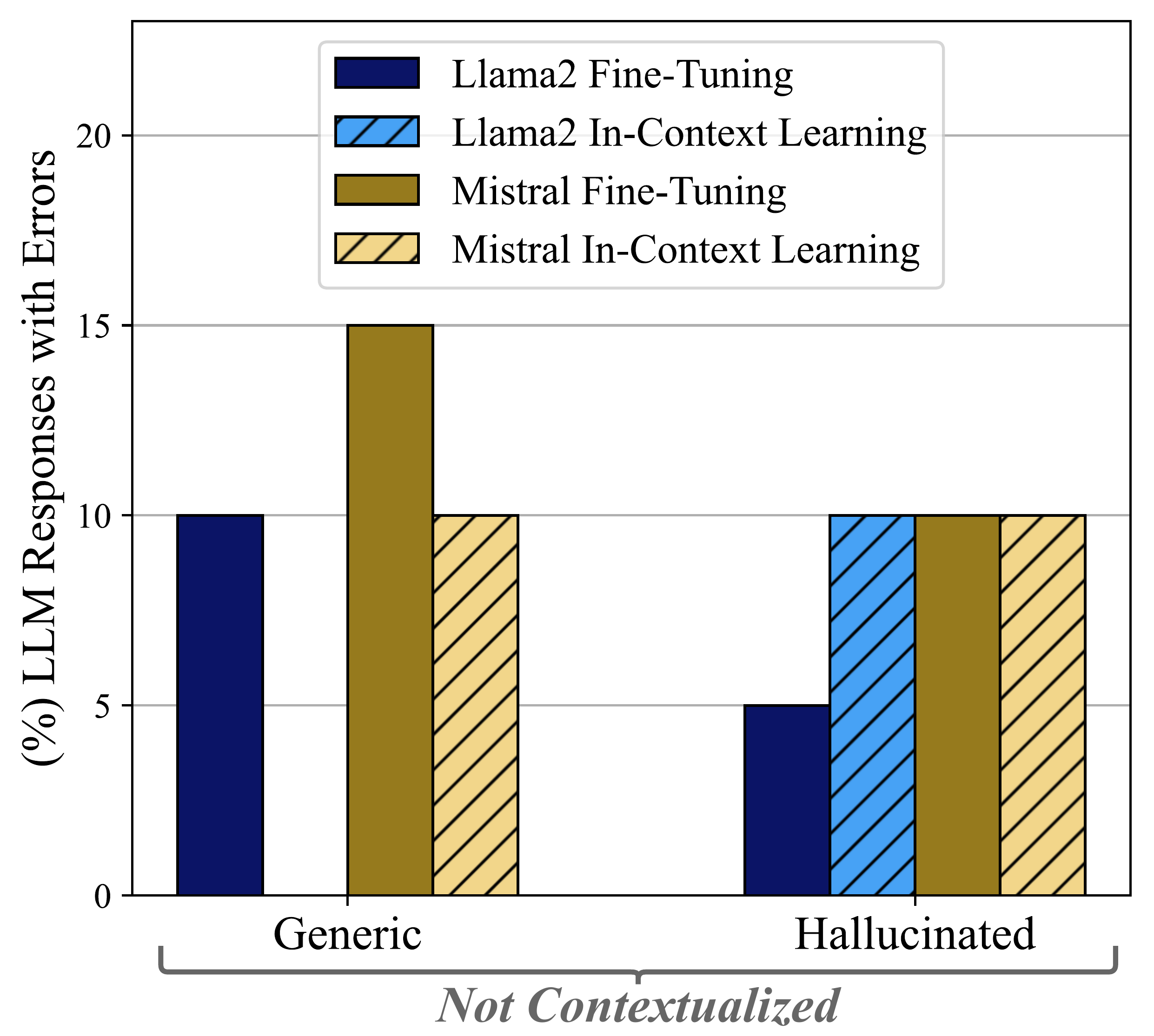}
    \caption{
        Percentage of LLM responses (y-axis) for each error type (\textit{Not Contextualized}) and their explanation (Generic, and Hallucinated) (x-axis), for \llama and \mistralp, adapted with In-Context Learning and Fine-Tuning in Question Answering (QA).
    }
    \label{fig:qa_hum_err}
\end{figure}

\subsection{Explaining Negative Human Judgments}
\label{sec:expl_hum_judg}
To better understand the shortcomings of the techniques, we investigate the motivations provided by the annotators to support their negative judgments. 
For each technique, we considered the scenario with gold external knowledge as the theoretical upper bound (except for ODDs where no external knowledge is required).
Following the original protocol, we consider two explanations for \textit{Not Contextualized} responses:
\begin{itemize}[noitemsep]
    \item \textbf{Generic}: the response is generic or does not contain any reference (implicit or explicit) to the dialogue history (ODD, KGD, TOD) or the gold knowledge (QA);
    \item \textbf{Hallucinated}: the response is inconsistent with the information contained in the dialogue history (ODD, KGD, TOD) or the gold knowledge (QA).
\end{itemize}
Regarding \textit{Not Appropriate} responses, the protocol has proposed one explanation (as an alternative to a free-form explanation):
\begin{itemize}
    \item \textbf{Incoherent}: the response is not coherent with the context.
\end{itemize}
To better characterize errors in TODs, we propose an additional explanation:
\begin{itemize}
    \item \textbf{Unhelpful}: the response candidate is not helpful in fulfilling the user's request.
\end{itemize}

In this section, we report the percentage of negatively judged responses with a certain explanation out of all the responses.

% ODD
\textbf{Open Domain Dialogue (ODD)}
In ODDs (Figure \ref{fig:odd_hum_err}), fine-tuning causes the generation of few generic responses, while for in-context learning none are present. Moreover, fine-tuned models generate around 30\% more hallucinated responses, and around 25\% more incoherent responses.

% KGD
\textbf{Knowledge-Grounded Dialogue (KGD)}
In KGDs (Figure \ref{fig:kgd_hum_err}), fine-tuning causes the generation of a few generic responses.
Regarding hallucinated responses, fine-tuning slightly reduces them for \llama but increases them for \mistralp.
Differently, fine-tuning slightly increases the incoherent responses for \llamap, but has no impact for \mistralp.

% TOD
\textbf{Task-Oriented Dialogue (TOD)}
For the TOD type (Figure \ref{fig:tod_hum_err}), while for \mistral fine-tuning has no impact on generic responses, it reduces generic responses by 15\% for \llamap.
For both models, fine-tuning reduces the number of hallucinated responses by 10\%,  and improves coherence by around 20\% both models. It further reduces unhelpful responses by 10\% for \llama.

% QA
\textbf{Question Answering (QA)}
For the QA type (Figure \ref{fig:qa_hum_err}), fine-tuned models generate more generic responses than models adapted with in-context learning.
Instead, fine-tuning results in fewer hallucinated responses for \llamap, although it has no effect for \mistralp.

\section{Conclusion}

We have conducted an extensive analysis on the efficacy of fine-tuning and in-context learning to adapt LLMs for different dialogue types. 
We have experimented with Retrieval-Augmented Generation (RAG) and gold knowledge to assess the impact of grounding the response generation on external knowledge. 
We have studied the models' performance using consistent criteria in both automatic (perplexity, explainability studies) and human evaluations.

Our study highlights the limitation of currently available automatic metrics and the necessity of conducting human evaluations to advance human-machine dialogue research, as the evaluations by human judges correlate poorly with automatic metrics. 
Furthermore, conducted human evaluations indicate that there is no universal best-technique for adapting LLMs to a dialogue type and the performance of each technique depends on the base LLM as well as the dialogue type. 
In addition, the correct incorporation of external knowledge depends on various factors such as the retriever accuracy, the representation of the knowledge, and the presence of noise (non-gold) documents, as it can be the least contributing element in the input vector according to explainability studies.

\section*{Limitations}
Due to the limited computational resources, we could experiment with 7B models, hampering us in validating our findings on larger models. 
Furthermore, the reproducibility of human evaluation results may be subject to variability, due to possible differences in the set of crowd workers.

\section*{Acknowledgments}
We acknowledge the support of the MUR PNRR project FAIR - Future AI Research (PE00000013) funded by the NextGenerationEU.

\bibliography{anthology,custom}

\begin{thebibliography}{68}
\expandafter\ifx\csname natexlab\endcsname\relax\def\natexlab#1{#1}\fi

\bibitem[{Baumgartner et~al.(2020)Baumgartner, Zannettou, Keegan, Squire, and Blackburn}]{baumgartner-2020-pushshift-reddit}
Jason Baumgartner, Savvas Zannettou, Brian Keegan, Megan Squire, and Jeremy Blackburn. 2020.
\newblock \href {https://doi.org/10.1609/icwsm.v14i1.7347} {The pushshift reddit dataset}.
\newblock \emph{Proceedings of the International AAAI Conference on Web and Social Media}, 14(1):830--839.

\bibitem[{Borgeaud et~al.(2022)Borgeaud, Mensch, Hoffmann, Cai, Rutherford, Millican, Van Den~Driessche, Lespiau, Damoc, Clark et~al.}]{borgeaud2022improving}
Sebastian Borgeaud, Arthur Mensch, Jordan Hoffmann, Trevor Cai, Eliza Rutherford, Katie Millican, George~Bm Van Den~Driessche, Jean-Baptiste Lespiau, Bogdan Damoc, Aidan Clark, et~al. 2022.
\newblock Improving language models by retrieving from trillions of tokens.
\newblock In \emph{International conference on machine learning}, pages 2206--2240. PMLR.

\bibitem[{Brown et~al.(2020)Brown, Mann, Ryder, Subbiah, Kaplan, Dhariwal, Neelakantan, Shyam, Sastry, Askell, Agarwal, Herbert-Voss, Krueger, Henighan, Child, Ramesh, Ziegler, Wu, Winter, Hesse, Chen, Sigler, Litwin, Gray, Chess, Clark, Berner, McCandlish, Radford, Sutskever, and Amodei}]{NEURIPS2020_1457c0d6}
Tom Brown, Benjamin Mann, Nick Ryder, Melanie Subbiah, Jared~D Kaplan, Prafulla Dhariwal, Arvind Neelakantan, Pranav Shyam, Girish Sastry, Amanda Askell, Sandhini Agarwal, Ariel Herbert-Voss, Gretchen Krueger, Tom Henighan, Rewon Child, Aditya Ramesh, Daniel Ziegler, Jeffrey Wu, Clemens Winter, Chris Hesse, Mark Chen, Eric Sigler, Mateusz Litwin, Scott Gray, Benjamin Chess, Jack Clark, Christopher Berner, Sam McCandlish, Alec Radford, Ilya Sutskever, and Dario Amodei. 2020.
\newblock \href {https://proceedings.neurips.cc/paper_files/paper/2020/file/1457c0d6bfcb4967418bfb8ac142f64a-Paper.pdf} {Language models are few-shot learners}.
\newblock In \emph{Advances in Neural Information Processing Systems}, volume~33, pages 1877--1901. Curran Associates, Inc.

\bibitem[{Chen et~al.(2023)Chen, Wu, and Li}]{chen-etal-2023-exploring-context}
Qinyu Chen, Wenhao Wu, and Sujian Li. 2023.
\newblock \href {https://doi.org/10.18653/v1/2023.findings-emnlp.675} {Exploring in-context learning for knowledge grounded dialog generation}.
\newblock In \emph{Findings of the Association for Computational Linguistics: EMNLP 2023}, pages 10071--10081, Singapore. Association for Computational Linguistics.

\bibitem[{Cho et~al.(2023)Cho, Seo, Jeong, and Park}]{cho-etal-2023-improving}
Sukmin Cho, Jeongyeon Seo, Soyeong Jeong, and Jong Park. 2023.
\newblock \href {https://doi.org/10.18653/v1/2023.findings-emnlp.207} {Improving zero-shot reader by reducing distractions from irrelevant documents in open-domain question answering}.
\newblock In \emph{Findings of the Association for Computational Linguistics: EMNLP 2023}, pages 3145--3157, Singapore. Association for Computational Linguistics.

\bibitem[{Dinan et~al.(2019)Dinan, Roller, Shuster, Fan, Auli, and Weston}]{dinan-2019-wizard}
Emily Dinan, Stephen Roller, Kurt Shuster, Angela Fan, Michael Auli, and Jason Weston. 2019.
\newblock {W}izard of {W}ikipedia: Knowledge-powered conversational agents.
\newblock In \emph{Proceedings of the International Conference on Learning Representations (ICLR)}.

\bibitem[{Ding et~al.(2024)Ding, Yang, Qiao, and Lin}]{ding-etal-2024-kmctod}
Zeyuan Ding, Zhihao Yang, Yinbo Qiao, and Hongfei Lin. 2024.
\newblock \href {https://doi.org/https://doi.org/10.1016/j.knosys.2024.111662} {Kmc-tod: Structure knowledge enhanced multi-copy network for task-oriented dialogue system}.
\newblock \emph{Knowledge-Based Systems}, 293:111662.

\bibitem[{Eric et~al.(2020)Eric, Goel, Paul, Sethi, Agarwal, Gao, Kumar, Goyal, Ku, and Hakkani-Tur}]{eric-etal-2020-multiwoz}
Mihail Eric, Rahul Goel, Shachi Paul, Abhishek Sethi, Sanchit Agarwal, Shuyang Gao, Adarsh Kumar, Anuj Goyal, Peter Ku, and Dilek Hakkani-Tur. 2020.
\newblock \href {https://aclanthology.org/2020.lrec-1.53} {{M}ulti{WOZ} 2.1: A consolidated multi-domain dialogue dataset with state corrections and state tracking baselines}.
\newblock In \emph{Proceedings of the Twelfth Language Resources and Evaluation Conference}, pages 422--428, Marseille, France. European Language Resources Association.

\bibitem[{Feng et~al.(2020)Feng, Wan, Gunasekara, Patel, Joshi, and Lastras}]{feng-etal-2020-doc2dial}
Song Feng, Hui Wan, Chulaka Gunasekara, Siva Patel, Sachindra Joshi, and Luis Lastras. 2020.
\newblock \href {https://doi.org/10.18653/v1/2020.emnlp-main.652} {doc2dial: A goal-oriented document-grounded dialogue dataset}.
\newblock In \emph{Proceedings of the 2020 Conference on Empirical Methods in Natural Language Processing (EMNLP)}, pages 8118--8128, Online. Association for Computational Linguistics.

\bibitem[{Fleiss(1971)}]{fleiss1971measuring}
Joseph~L Fleiss. 1971.
\newblock Measuring nominal scale agreement among many raters.
\newblock \emph{Psychological bulletin}, 76(5):378.

\bibitem[{Godfrey et~al.(1992)Godfrey, Holliman, and McDaniel}]{godfrey-1992-switchboard}
J.J. Godfrey, E.C. Holliman, and J.~McDaniel. 1992.
\newblock \href {https://doi.org/10.1109/ICASSP.1992.225858} {Switchboard: telephone speech corpus for research and development}.
\newblock In \emph{[Proceedings] ICASSP-92: 1992 IEEE International Conference on Acoustics, Speech, and Signal Processing}, volume~1, pages 517--520 vol.1.

\bibitem[{Gopalakrishnan et~al.(2019)Gopalakrishnan, Hedayatnia, Chen, Gottardi, Kwatra, Venkatesh, Gabriel, and Hakkani-Tür}]{gopalakrishnan-2019-topical}
Karthik Gopalakrishnan, Behnam Hedayatnia, Qinlang Chen, Anna Gottardi, Sanjeev Kwatra, Anu Venkatesh, Raefer Gabriel, and Dilek Hakkani-Tür. 2019.
\newblock \href {https://doi.org/10.21437/Interspeech.2019-3079} {{Topical-Chat: Towards Knowledge-Grounded Open-Domain Conversations}}.
\newblock In \emph{Proc. Interspeech 2019}, pages 1891--1895.

\bibitem[{Han et~al.(2023)Han, Jo, Nam, Yoon, Kwon, Rho, On, Yoo, and Kim}]{han-etal-2023-efficient}
Gunsoo Han, Daejin Jo, Daniel Nam, Eunseop Yoon, Taehwan Kwon, Seungeun Rho, Kyoung-Woon On, Chang Yoo, and Sungwoong Kim. 2023.
\newblock \href {https://doi.org/10.18653/v1/2023.findings-emnlp.177} {Efficient latent variable modeling for knowledge-grounded dialogue generation}.
\newblock In \emph{Findings of the Association for Computational Linguistics: EMNLP 2023}, pages 2683--2702, Singapore. Association for Computational Linguistics.

\bibitem[{He et~al.(2024)He, Lu, Bao, Wang, Wu, Niu, and Wang}]{he-etal-2023-dstc9winner}
Huang He, Hua Lu, Siqi Bao, Fan Wang, Hua Wu, Zheng-Yu Niu, and Haifeng Wang. 2024.
\newblock \href {https://doi.org/10.1109/TASLP.2023.3301222} {Learning to select external knowledge with multi-scale negative sampling}.
\newblock \emph{IEEE/ACM Transactions on Audio, Speech, and Language Processing}, 32:714--720.

\bibitem[{Hedayatnia et~al.(2020)Hedayatnia, Gopalakrishnan, Kim, Liu, Eric, and Hakkani-Tur}]{hedayatnia-etal-2020-policy}
Behnam Hedayatnia, Karthik Gopalakrishnan, Seokhwan Kim, Yang Liu, Mihail Eric, and Dilek Hakkani-Tur. 2020.
\newblock \href {https://doi.org/10.18653/v1/2020.inlg-1.46} {Policy-driven neural response generation for knowledge-grounded dialog systems}.
\newblock In \emph{Proceedings of the 13th International Conference on Natural Language Generation}, pages 412--421, Dublin, Ireland. Association for Computational Linguistics.

\bibitem[{Hosseini-Asl et~al.(2020{\natexlab{a}})Hosseini-Asl, McCann, Wu, Yavuz, and Socher}]{hosseini-asl-2020-simpletod}
Ehsan Hosseini-Asl, Bryan McCann, Chien-Sheng Wu, Semih Yavuz, and Richard Socher. 2020{\natexlab{a}}.
\newblock \href {https://proceedings.neurips.cc/paper_files/paper/2020/file/e946209592563be0f01c844ab2170f0c-Paper.pdf} {A simple language model for task-oriented dialogue}.
\newblock In \emph{Advances in Neural Information Processing Systems}, volume~33, pages 20179--20191. Curran Associates, Inc.

\bibitem[{Hosseini-Asl et~al.(2020{\natexlab{b}})Hosseini-Asl, McCann, Wu, Yavuz, and Socher}]{hosseini-asl-etal-2020-simpletod}
Ehsan Hosseini-Asl, Bryan McCann, Chien-Sheng Wu, Semih Yavuz, and Richard Socher. 2020{\natexlab{b}}.
\newblock \href {https://proceedings.neurips.cc/paper_files/paper/2020/file/e946209592563be0f01c844ab2170f0c-Paper.pdf} {A simple language model for task-oriented dialogue}.
\newblock In \emph{Advances in Neural Information Processing Systems}, volume~33, pages 20179--20191. Curran Associates, Inc.

\bibitem[{Hu et~al.(2021)Hu, Shen, Wallis, Allen-Zhu, Li, Wang, Wang, and Chen}]{hu-2021-lora}
Edward~J Hu, Yelong Shen, Phillip Wallis, Zeyuan Allen-Zhu, Yuanzhi Li, Shean Wang, Lu~Wang, and Weizhu Chen. 2021.
\newblock Lora: Low-rank adaptation of large language models.
\newblock \emph{arXiv preprint arXiv:2106.09685}.

\bibitem[{Huang et~al.(2023)Huang, Fu, Liu, Wang, Ko, Zhang, and Tang}]{huang-etal-2023-learning}
Qiushi Huang, Shuai Fu, Xubo Liu, Wenwu Wang, Tom Ko, Yu~Zhang, and Lilian Tang. 2023.
\newblock \href {https://doi.org/10.18653/v1/2023.emnlp-main.154} {Learning retrieval augmentation for personalized dialogue generation}.
\newblock In \emph{Proceedings of the 2023 Conference on Empirical Methods in Natural Language Processing}, pages 2523--2540, Singapore. Association for Computational Linguistics.

\bibitem[{Hude{\v{c}}ek and Dusek(2023)}]{hudecek-dusek-2023-large}
Vojt{\v{e}}ch Hude{\v{c}}ek and Ondrej Dusek. 2023.
\newblock \href {https://doi.org/10.18653/v1/2023.sigdial-1.21} {Are large language models all you need for task-oriented dialogue?}
\newblock In \emph{Proceedings of the 24th Annual Meeting of the Special Interest Group on Discourse and Dialogue}, pages 216--228, Prague, Czechia. Association for Computational Linguistics.

\bibitem[{Izacard and Grave(2021)}]{izacard-grave-2021-leveraging}
Gautier Izacard and Edouard Grave. 2021.
\newblock \href {https://doi.org/10.18653/v1/2021.eacl-main.74} {Leveraging passage retrieval with generative models for open domain question answering}.
\newblock In \emph{Proceedings of the 16th Conference of the European Chapter of the Association for Computational Linguistics: Main Volume}, pages 874--880, Online. Association for Computational Linguistics.

\bibitem[{Jiang et~al.(2023)Jiang, Sablayrolles, Mensch, Bamford, Chaplot, de~las Casas, Bressand, Lengyel, Lample, Saulnier, Lavaud, Lachaux, Stock, Scao, Lavril, Wang, Lacroix, and Sayed}]{jiang2023mistral}
Albert~Q. Jiang, Alexandre Sablayrolles, Arthur Mensch, Chris Bamford, Devendra~Singh Chaplot, Diego de~las Casas, Florian Bressand, Gianna Lengyel, Guillaume Lample, Lucile Saulnier, Lélio~Renard Lavaud, Marie-Anne Lachaux, Pierre Stock, Teven~Le Scao, Thibaut Lavril, Thomas Wang, Timothée Lacroix, and William~El Sayed. 2023.
\newblock \href {http://arxiv.org/abs/2310.06825} {Mistral 7b}.

\bibitem[{Karpukhin et~al.(2020)Karpukhin, Oguz, Min, Lewis, Wu, Edunov, Chen, and Yih}]{karpukhin-etal-2020-dense}
Vladimir Karpukhin, Barlas Oguz, Sewon Min, Patrick Lewis, Ledell Wu, Sergey Edunov, Danqi Chen, and Wen-tau Yih. 2020.
\newblock \href {https://doi.org/10.18653/v1/2020.emnlp-main.550} {Dense passage retrieval for open-domain question answering}.
\newblock In \emph{Proceedings of the 2020 Conference on Empirical Methods in Natural Language Processing (EMNLP)}, pages 6769--6781, Online. Association for Computational Linguistics.

\bibitem[{Kasahara et~al.(2022)Kasahara, Kawahara, Tung, Li, Shinzato, and Sato}]{kasahara-etal-2022-building}
Tomohito Kasahara, Daisuke Kawahara, Nguyen Tung, Shengzhe Li, Kenta Shinzato, and Toshinori Sato. 2022.
\newblock \href {https://doi.org/10.18653/v1/2022.naacl-srw.13} {Building a personalized dialogue system with prompt-tuning}.
\newblock In \emph{Proceedings of the 2022 Conference of the North American Chapter of the Association for Computational Linguistics: Human Language Technologies: Student Research Workshop}, pages 96--105, Hybrid: Seattle, Washington + Online. Association for Computational Linguistics.

\bibitem[{Kim et~al.(2020)Kim, Eric, Gopalakrishnan, Hedayatnia, Liu, and Hakkani-Tur}]{kim-etal-2020-beyond}
Seokhwan Kim, Mihail Eric, Karthik Gopalakrishnan, Behnam Hedayatnia, Yang Liu, and Dilek Hakkani-Tur. 2020.
\newblock \href {https://doi.org/10.18653/v1/2020.sigdial-1.35} {Beyond domain {API}s: Task-oriented conversational modeling with unstructured knowledge access}.
\newblock In \emph{Proceedings of the 21th Annual Meeting of the Special Interest Group on Discourse and Dialogue}, pages 278--289, 1st virtual meeting. Association for Computational Linguistics.

\bibitem[{Kim et~al.(2021)Kim, Liu, Jin, Papangelis, Gopalakrishnan, Hedayatnia, and Hakkani-Tür}]{kim-2021-dstc10}
Seokhwan Kim, Yang Liu, Di~Jin, Alexandros Papangelis, Karthik Gopalakrishnan, Behnam Hedayatnia, and Dilek Hakkani-Tür. 2021.
\newblock \href {https://doi.org/10.1109/ASRU51503.2021.9688274} {“how robust r u?”: Evaluating task-oriented dialogue systems on spoken conversations}.
\newblock In \emph{2021 IEEE Automatic Speech Recognition and Understanding Workshop (ASRU)}, pages 1147--1154.

\bibitem[{Ko{\v{c}}isk{\'y} et~al.(2018)Ko{\v{c}}isk{\'y}, Schwarz, Blunsom, Dyer, Hermann, Melis, and Grefenstette}]{kocisky-etal-2018-narrativeqa}
Tom{\'a}{\v{s}} Ko{\v{c}}isk{\'y}, Jonathan Schwarz, Phil Blunsom, Chris Dyer, Karl~Moritz Hermann, G{\'a}bor Melis, and Edward Grefenstette. 2018.
\newblock \href {https://doi.org/10.1162/tacl_a_00023} {The {N}arrative{QA} reading comprehension challenge}.
\newblock \emph{Transactions of the Association for Computational Linguistics}, 6:317--328.

\bibitem[{Komeili et~al.(2022)Komeili, Shuster, and Weston}]{komeili-etal-2022-internet}
Mojtaba Komeili, Kurt Shuster, and Jason Weston. 2022.
\newblock \href {https://doi.org/10.18653/v1/2022.acl-long.579} {{I}nternet-augmented dialogue generation}.
\newblock In \emph{Proceedings of the 60th Annual Meeting of the Association for Computational Linguistics (Volume 1: Long Papers)}, pages 8460--8478, Dublin, Ireland. Association for Computational Linguistics.

\bibitem[{Kulh{\'a}nek et~al.(2021)Kulh{\'a}nek, Hude{\v{c}}ek, Nekvinda, and Du{\v{s}}ek}]{kulhanek-etal-2021-augpt}
Jon{\'a}{\v{s}} Kulh{\'a}nek, Vojt{\v{e}}ch Hude{\v{c}}ek, Tom{\'a}{\v{s}} Nekvinda, and Ond{\v{r}}ej Du{\v{s}}ek. 2021.
\newblock \href {https://doi.org/10.18653/v1/2021.nlp4convai-1.19} {{AuGPT}: Auxiliary tasks and data augmentation for end-to-end dialogue with pre-trained language models}.
\newblock In \emph{Proceedings of the 3rd Workshop on Natural Language Processing for Conversational AI}, pages 198--210, Online. Association for Computational Linguistics.

\bibitem[{Kwiatkowski et~al.(2019)Kwiatkowski, Palomaki, Redfield, Collins, Parikh, Alberti, Epstein, Polosukhin, Devlin, Lee, Toutanova, Jones, Kelcey, Chang, Dai, Uszkoreit, Le, and Petrov}]{kwiatkowski-etal-2019-natural}
Tom Kwiatkowski, Jennimaria Palomaki, Olivia Redfield, Michael Collins, Ankur Parikh, Chris Alberti, Danielle Epstein, Illia Polosukhin, Jacob Devlin, Kenton Lee, Kristina Toutanova, Llion Jones, Matthew Kelcey, Ming-Wei Chang, Andrew~M. Dai, Jakob Uszkoreit, Quoc Le, and Slav Petrov. 2019.
\newblock \href {https://doi.org/10.1162/tacl_a_00276} {Natural questions: A benchmark for question answering research}.
\newblock \emph{Transactions of the Association for Computational Linguistics}, 7:452--466.

\bibitem[{Lee et~al.(2019)Lee, Chang, and Toutanova}]{lee-etal-2019-latent}
Kenton Lee, Ming-Wei Chang, and Kristina Toutanova. 2019.
\newblock \href {https://doi.org/10.18653/v1/P19-1612} {Latent retrieval for weakly supervised open domain question answering}.
\newblock In \emph{Proceedings of the 57th Annual Meeting of the Association for Computational Linguistics}, pages 6086--6096, Florence, Italy. Association for Computational Linguistics.

\bibitem[{Levine et~al.(2022)Levine, Ram, Jannai, Lenz, Shalev-Shwartz, Shashua, Leyton-Brown, and Shoham}]{levine2022huge}
Yoav Levine, Ori Ram, Daniel Jannai, Barak Lenz, Shai Shalev-Shwartz, Amnon Shashua, Kevin Leyton-Brown, and Yoav Shoham. 2022.
\newblock \href {https://openreview.net/forum?id=z3Bxu8xNJaF} {Huge frozen language models as readers for open-domain question answering}.
\newblock In \emph{ICML 2022 Workshop on Knowledge Retrieval and Language Models}.

\bibitem[{Lewis et~al.(2020)Lewis, Perez, Piktus, Petroni, Karpukhin, Goyal, K\"{u}ttler, Lewis, Yih, Rockt\"{a}schel, Riedel, and Kiela}]{lewis-2020-retrieval}
Patrick Lewis, Ethan Perez, Aleksandra Piktus, Fabio Petroni, Vladimir Karpukhin, Naman Goyal, Heinrich K\"{u}ttler, Mike Lewis, Wen-tau Yih, Tim Rockt\"{a}schel, Sebastian Riedel, and Douwe Kiela. 2020.
\newblock \href {https://proceedings.neurips.cc/paper_files/paper/2020/file/6b493230205f780e1bc26945df7481e5-Paper.pdf} {Retrieval-augmented generation for knowledge-intensive nlp tasks}.
\newblock In \emph{Advances in Neural Information Processing Systems}, volume~33, pages 9459--9474. Curran Associates, Inc.

\bibitem[{Li et~al.(2017)Li, Su, Shen, Li, Cao, and Niu}]{li-etal-2017-dailydialog}
Yanran Li, Hui Su, Xiaoyu Shen, Wenjie Li, Ziqiang Cao, and Shuzi Niu. 2017.
\newblock \href {https://aclanthology.org/I17-1099} {{D}aily{D}ialog: A manually labelled multi-turn dialogue dataset}.
\newblock In \emph{Proceedings of the Eighth International Joint Conference on Natural Language Processing (Volume 1: Long Papers)}, pages 986--995, Taipei, Taiwan. Asian Federation of Natural Language Processing.

\bibitem[{Lin(2004)}]{lin-2004-rouge}
Chin-Yew Lin. 2004.
\newblock \href {https://aclanthology.org/W04-1013} {{ROUGE}: A package for automatic evaluation of summaries}.
\newblock In \emph{Text Summarization Branches Out}, pages 74--81, Barcelona, Spain. Association for Computational Linguistics.

\bibitem[{Lin and Chen(2023)}]{lin-chen-2023-llm}
Yen-Ting Lin and Yun-Nung Chen. 2023.
\newblock \href {https://doi.org/10.18653/v1/2023.nlp4convai-1.5} {{LLM}-eval: Unified multi-dimensional automatic evaluation for open-domain conversations with large language models}.
\newblock In \emph{Proceedings of the 5th Workshop on NLP for Conversational AI (NLP4ConvAI 2023)}, pages 47--58, Toronto, Canada. Association for Computational Linguistics.

\bibitem[{Liu et~al.(2016)Liu, Lowe, Serban, Noseworthy, Charlin, and Pineau}]{liu-etal-2016-evaluate}
Chia-Wei Liu, Ryan Lowe, Iulian Serban, Mike Noseworthy, Laurent Charlin, and Joelle Pineau. 2016.
\newblock \href {https://doi.org/10.18653/v1/D16-1230} {How {NOT} to evaluate your dialogue system: An empirical study of unsupervised evaluation metrics for dialogue response generation}.
\newblock In \emph{Proceedings of the 2016 Conference on Empirical Methods in Natural Language Processing}, pages 2122--2132, Austin, Texas. Association for Computational Linguistics.

\bibitem[{Loshchilov and Hutter(2017)}]{loshchilov2017decoupled}
Ilya Loshchilov and Frank Hutter. 2017.
\newblock Decoupled weight decay regularization.
\newblock \emph{arXiv preprint arXiv:1711.05101}.

\bibitem[{Meade et~al.(2023)Meade, Gella, Hazarika, Gupta, Jin, Reddy, Liu, and Hakkani-Tur}]{meade-etal-2023-using}
Nicholas Meade, Spandana Gella, Devamanyu Hazarika, Prakhar Gupta, Di~Jin, Siva Reddy, Yang Liu, and Dilek Hakkani-Tur. 2023.
\newblock \href {https://doi.org/10.18653/v1/2023.findings-emnlp.796} {Using in-context learning to improve dialogue safety}.
\newblock In \emph{Findings of the Association for Computational Linguistics: EMNLP 2023}, pages 11882--11910, Singapore. Association for Computational Linguistics.

\bibitem[{Mousavi et~al.(2023)Mousavi, Caldarella, and Riccardi}]{mousavi-etal-2023-response}
Seyed~Mahed Mousavi, Simone Caldarella, and Giuseppe Riccardi. 2023.
\newblock \href {https://doi.org/10.18653/v1/2023.nlp4convai-1.1} {Response generation in longitudinal dialogues: Which knowledge representation helps?}
\newblock In \emph{Proceedings of the 5th Workshop on NLP for Conversational AI (NLP4ConvAI 2023)}, pages 1--11, Toronto, Canada. Association for Computational Linguistics.

\bibitem[{Mousavi et~al.(2024)Mousavi, Roccabruna, Alghisi, Rizzoli, Ravanelli, and Riccardi}]{mousavi2024llms}
Seyed~Mahed Mousavi, Gabriel Roccabruna, Simone Alghisi, Massimo Rizzoli, Mirco Ravanelli, and Giuseppe Riccardi. 2024.
\newblock \href {http://arxiv.org/abs/2401.02297} {Are llms robust for spoken dialogues?}

\bibitem[{Mousavi et~al.(2022)Mousavi, Roccabruna, Lorandi, Caldarella, and Riccardi}]{mousavi-etal-2022-evaluation}
Seyed~Mahed Mousavi, Gabriel Roccabruna, Michela Lorandi, Simone Caldarella, and Giuseppe Riccardi. 2022.
\newblock \href {https://doi.org/10.18653/v1/2022.gem-1.12} {Evaluation of response generation models: Shouldn{'}t it be shareable and replicable?}
\newblock In \emph{Proceedings of the 2nd Workshop on Natural Language Generation, Evaluation, and Metrics (GEM)}, pages 136--147, Abu Dhabi, United Arab Emirates (Hybrid). Association for Computational Linguistics.

\bibitem[{Papineni et~al.(2002)Papineni, Roukos, Ward, and Zhu}]{papineni-etal-2002-bleu}
Kishore Papineni, Salim Roukos, Todd Ward, and Wei-Jing Zhu. 2002.
\newblock \href {https://doi.org/10.3115/1073083.1073135} {{B}leu: a method for automatic evaluation of machine translation}.
\newblock In \emph{Proceedings of the 40th Annual Meeting of the Association for Computational Linguistics}, pages 311--318, Philadelphia, Pennsylvania, USA. Association for Computational Linguistics.

\bibitem[{Qian et~al.(2023)Qian, Zhang, and Liu}]{qian-etal-2023-harnessing}
Yushan Qian, Weinan Zhang, and Ting Liu. 2023.
\newblock \href {https://doi.org/10.18653/v1/2023.findings-emnlp.433} {Harnessing the power of large language models for empathetic response generation: Empirical investigations and improvements}.
\newblock In \emph{Findings of the Association for Computational Linguistics: EMNLP 2023}, pages 6516--6528, Singapore. Association for Computational Linguistics.

\bibitem[{Qin et~al.(2023)Qin, Zhang, Liang, Wang, and Yang}]{qin-etal-2023-well}
Lang Qin, Yao Zhang, Hongru Liang, Jun Wang, and Zhenglu Yang. 2023.
\newblock \href {https://doi.org/10.18653/v1/2023.emnlp-main.285} {Well begun is half done: Generator-agnostic knowledge pre-selection for knowledge-grounded dialogue}.
\newblock In \emph{Proceedings of the 2023 Conference on Empirical Methods in Natural Language Processing}, pages 4696--4709, Singapore. Association for Computational Linguistics.

\bibitem[{Qu et~al.(2020)Qu, Yang, Chen, Qiu, Croft, and Iyyer}]{qu-2020-open}
Chen Qu, Liu Yang, Cen Chen, Minghui Qiu, W.~Bruce Croft, and Mohit Iyyer. 2020.
\newblock \href {https://doi.org/10.1145/3397271.3401110} {Open-retrieval conversational question answering}.
\newblock In \emph{Proceedings of the 43rd International ACM SIGIR Conference on Research and Development in Information Retrieval}, SIGIR '20, page 539–548, New York, NY, USA. Association for Computing Machinery.

\bibitem[{Rajpurkar et~al.(2018)Rajpurkar, Jia, and Liang}]{rajpurkar-etal-2018-know}
Pranav Rajpurkar, Robin Jia, and Percy Liang. 2018.
\newblock \href {https://doi.org/10.18653/v1/P18-2124} {Know what you don{'}t know: Unanswerable questions for {SQ}u{AD}}.
\newblock In \emph{Proceedings of the 56th Annual Meeting of the Association for Computational Linguistics (Volume 2: Short Papers)}, pages 784--789, Melbourne, Australia. Association for Computational Linguistics.

\bibitem[{Rajpurkar et~al.(2016)Rajpurkar, Zhang, Lopyrev, and Liang}]{rajpurkar-etal-2016-squad}
Pranav Rajpurkar, Jian Zhang, Konstantin Lopyrev, and Percy Liang. 2016.
\newblock \href {https://doi.org/10.18653/v1/D16-1264} {{SQ}u{AD}: 100,000+ questions for machine comprehension of text}.
\newblock In \emph{Proceedings of the 2016 Conference on Empirical Methods in Natural Language Processing}, pages 2383--2392, Austin, Texas. Association for Computational Linguistics.

\bibitem[{Raposo et~al.(2023)Raposo, Coheur, and Martins}]{raposo-etal-2023-prompting}
Gon{\c{c}}alo Raposo, Luisa Coheur, and Bruno Martins. 2023.
\newblock \href {https://doi.org/10.18653/v1/2023.sigdial-1.37} {Prompting, retrieval, training: An exploration of different approaches for task-oriented dialogue generation}.
\newblock In \emph{Proceedings of the 24th Annual Meeting of the Special Interest Group on Discourse and Dialogue}, pages 400--412, Prague, Czechia. Association for Computational Linguistics.

\bibitem[{Rashkin et~al.(2019)Rashkin, Smith, Li, and Boureau}]{rashkin-etal-2019-towards}
Hannah Rashkin, Eric~Michael Smith, Margaret Li, and Y-Lan Boureau. 2019.
\newblock \href {https://doi.org/10.18653/v1/P19-1534} {Towards empathetic open-domain conversation models: A new benchmark and dataset}.
\newblock In \emph{Proceedings of the 57th Annual Meeting of the Association for Computational Linguistics}, pages 5370--5381, Florence, Italy. Association for Computational Linguistics.

\bibitem[{Sai et~al.(2022)Sai, Mohankumar, and Khapra}]{sai2022surveyevalnlg}
Ananya~B. Sai, Akash~Kumar Mohankumar, and Mitesh~M. Khapra. 2022.
\newblock \href {https://doi.org/10.1145/3485766} {A survey of evaluation metrics used for nlg systems}.
\newblock \emph{ACM Comput. Surv.}, 55(2).

\bibitem[{Sarti et~al.(2023)Sarti, Feldhus, Sickert, and van~der Wal}]{sarti-etal-2023-inseq}
Gabriele Sarti, Nils Feldhus, Ludwig Sickert, and Oskar van~der Wal. 2023.
\newblock \href {https://doi.org/10.18653/v1/2023.acl-demo.40} {Inseq: An interpretability toolkit for sequence generation models}.
\newblock In \emph{Proceedings of the 61st Annual Meeting of the Association for Computational Linguistics (Volume 3: System Demonstrations)}, pages 421--435, Toronto, Canada. Association for Computational Linguistics.

\bibitem[{Shuster et~al.(2021)Shuster, Poff, Chen, Kiela, and Weston}]{shuster-etal-2021-retrieval-augmentation}
Kurt Shuster, Spencer Poff, Moya Chen, Douwe Kiela, and Jason Weston. 2021.
\newblock \href {https://doi.org/10.18653/v1/2021.findings-emnlp.320} {Retrieval augmentation reduces hallucination in conversation}.
\newblock In \emph{Findings of the Association for Computational Linguistics: EMNLP 2021}, pages 3784--3803, Punta Cana, Dominican Republic. Association for Computational Linguistics.

\bibitem[{Sun et~al.(2023)Sun, Ren, and Ren}]{sun-etal-2023-generative}
Weiwei Sun, Pengjie Ren, and Zhaochun Ren. 2023.
\newblock \href {https://doi.org/10.18653/v1/2023.findings-eacl.155} {Generative knowledge selection for knowledge-grounded dialogues}.
\newblock In \emph{Findings of the Association for Computational Linguistics: EACL 2023}, pages 2077--2088, Dubrovnik, Croatia. Association for Computational Linguistics.

\bibitem[{Thulke et~al.(2024)Thulke, Daheim, Dugast, and Ney}]{thulke-2024-tod-dstc9-10}
David Thulke, Nico Daheim, Christian Dugast, and Hermann Ney. 2024.
\newblock \href {https://doi.org/10.1109/TASLP.2023.3267832} {Task-oriented document-grounded dialog systems by hltpr@rwth for dstc9 and dstc10}.
\newblock \emph{IEEE/ACM Transactions on Audio, Speech, and Language Processing}, 32:733--741.

\bibitem[{Tiedemann(2009)}]{tiedemann-2009-opensubtitles}
Jörg Tiedemann. 2009.
\newblock \href {https://doi.org/10.1075/cilt.309.19tie} {\emph{News from OPUS—A Collection of Multilingual Parallel Corpora with Tools and Interfaces}}, volume~5, pages 237--248.

\bibitem[{Touvron et~al.(2023)Touvron, Martin, Stone, Albert, Almahairi, Babaei, Bashlykov, Batra, Bhargava, Bhosale, Bikel, Blecher, Ferrer, Chen, Cucurull, Esiobu, Fernandes, Fu, Fu, Fuller, Gao, Goswami, Goyal, Hartshorn, Hosseini, Hou, Inan, Kardas, Kerkez, Khabsa, Kloumann, Korenev, Koura, Lachaux, Lavril, Lee, Liskovich, Lu, Mao, Martinet, Mihaylov, Mishra, Molybog, Nie, Poulton, Reizenstein, Rungta, Saladi, Schelten, Silva, Smith, Subramanian, Tan, Tang, Taylor, Williams, Kuan, Xu, Yan, Zarov, Zhang, Fan, Kambadur, Narang, Rodriguez, Stojnic, Edunov, and Scialom}]{touvron2023llama}
Hugo Touvron, Louis Martin, Kevin Stone, Peter Albert, Amjad Almahairi, Yasmine Babaei, Nikolay Bashlykov, Soumya Batra, Prajjwal Bhargava, Shruti Bhosale, Dan Bikel, Lukas Blecher, Cristian~Canton Ferrer, Moya Chen, Guillem Cucurull, David Esiobu, Jude Fernandes, Jeremy Fu, Wenyin Fu, Brian Fuller, Cynthia Gao, Vedanuj Goswami, Naman Goyal, Anthony Hartshorn, Saghar Hosseini, Rui Hou, Hakan Inan, Marcin Kardas, Viktor Kerkez, Madian Khabsa, Isabel Kloumann, Artem Korenev, Punit~Singh Koura, Marie-Anne Lachaux, Thibaut Lavril, Jenya Lee, Diana Liskovich, Yinghai Lu, Yuning Mao, Xavier Martinet, Todor Mihaylov, Pushkar Mishra, Igor Molybog, Yixin Nie, Andrew Poulton, Jeremy Reizenstein, Rashi Rungta, Kalyan Saladi, Alan Schelten, Ruan Silva, Eric~Michael Smith, Ranjan Subramanian, Xiaoqing~Ellen Tan, Binh Tang, Ross Taylor, Adina Williams, Jian~Xiang Kuan, Puxin Xu, Zheng Yan, Iliyan Zarov, Yuchen Zhang, Angela Fan, Melanie Kambadur, Sharan Narang, Aurelien Rodriguez, Robert Stojnic, Sergey Edunov, and Thomas
  Scialom. 2023.
\newblock \href {http://arxiv.org/abs/2307.09288} {Llama 2: Open foundation and fine-tuned chat models}.

\bibitem[{Wang et~al.(2022)Wang, Zhang, Guo, Dai, Chen, and Luo}]{wang-2022-todnlg}
Weizhi Wang, Zhirui Zhang, Junliang Guo, Yinpei Dai, Boxing Chen, and Weihua Luo. 2022.
\newblock \href {https://doi.org/10.1145/3477495.3531920} {Task-oriented dialogue system as natural language generation}.
\newblock In \emph{Proceedings of the 45th International ACM SIGIR Conference on Research and Development in Information Retrieval}, SIGIR '22, page 2698–2703, New York, NY, USA. Association for Computing Machinery.

\bibitem[{Wolf et~al.(2019)Wolf, Sanh, Chaumond, and Delangue}]{wolf-2019-transfertransfo}
Thomas Wolf, Victor Sanh, Julien Chaumond, and Clement Delangue. 2019.
\newblock Transfertransfo: A transfer learning approach for neural network based conversational agents.
\newblock \emph{arXiv preprint arXiv:1901.08149}.

\bibitem[{Xu et~al.(2022{\natexlab{a}})Xu, Szlam, and Weston}]{xu-etal-2022-beyond}
Jing Xu, Arthur Szlam, and Jason Weston. 2022{\natexlab{a}}.
\newblock \href {https://doi.org/10.18653/v1/2022.acl-long.356} {Beyond goldfish memory: Long-term open-domain conversation}.
\newblock In \emph{Proceedings of the 60th Annual Meeting of the Association for Computational Linguistics (Volume 1: Long Papers)}, pages 5180--5197, Dublin, Ireland. Association for Computational Linguistics.

\bibitem[{Xu et~al.(2022{\natexlab{b}})Xu, Gou, Wu, Niu, Wu, Wang, and Wang}]{xu-etal-2022-long}
Xinchao Xu, Zhibin Gou, Wenquan Wu, Zheng-Yu Niu, Hua Wu, Haifeng Wang, and Shihang Wang. 2022{\natexlab{b}}.
\newblock \href {https://doi.org/10.18653/v1/2022.findings-acl.207} {Long time no see! open-domain conversation with long-term persona memory}.
\newblock In \emph{Findings of the Association for Computational Linguistics: ACL 2022}, pages 2639--2650, Dublin, Ireland. Association for Computational Linguistics.

\bibitem[{Yang et~al.(2023)Yang, Huang, Liu, and Gao}]{yang-etal-2023-graph}
Yizhe Yang, Heyan Huang, Yuhang Liu, and Yang Gao. 2023.
\newblock \href {https://doi.org/10.18653/v1/2023.emnlp-main.982} {Graph vs. sequence: An empirical study on knowledge forms for knowledge-grounded dialogue}.
\newblock In \emph{Proceedings of the 2023 Conference on Empirical Methods in Natural Language Processing}, pages 15846--15858, Singapore. Association for Computational Linguistics.

\bibitem[{Yang et~al.(2018)Yang, Qi, Zhang, Bengio, Cohen, Salakhutdinov, and Manning}]{yang-etal-2018-hotpotqa}
Zhilin Yang, Peng Qi, Saizheng Zhang, Yoshua Bengio, William Cohen, Ruslan Salakhutdinov, and Christopher~D. Manning. 2018.
\newblock \href {https://doi.org/10.18653/v1/D18-1259} {{H}otpot{QA}: A dataset for diverse, explainable multi-hop question answering}.
\newblock In \emph{Proceedings of the 2018 Conference on Empirical Methods in Natural Language Processing}, pages 2369--2380, Brussels, Belgium. Association for Computational Linguistics.

\bibitem[{Zhang et~al.(2023)Zhang, Chen, Xu, Cao, Chen, Cohn, and Fang}]{zhang-etal-2023-survey-efficient}
Qin Zhang, Shangsi Chen, Dongkuan Xu, Qingqing Cao, Xiaojun Chen, Trevor Cohn, and Meng Fang. 2023.
\newblock \href {https://doi.org/10.18653/v1/2023.acl-long.808} {A survey for efficient open domain question answering}.
\newblock In \emph{Proceedings of the 61st Annual Meeting of the Association for Computational Linguistics (Volume 1: Long Papers)}, pages 14447--14465, Toronto, Canada. Association for Computational Linguistics.

\bibitem[{Zhang et~al.(2018)Zhang, Dinan, Urbanek, Szlam, Kiela, and Weston}]{zhang-etal-2018-personalizing}
Saizheng Zhang, Emily Dinan, Jack Urbanek, Arthur Szlam, Douwe Kiela, and Jason Weston. 2018.
\newblock \href {https://doi.org/10.18653/v1/P18-1205} {Personalizing dialogue agents: {I} have a dog, do you have pets too?}
\newblock In \emph{Proceedings of the 56th Annual Meeting of the Association for Computational Linguistics (Volume 1: Long Papers)}, pages 2204--2213, Melbourne, Australia. Association for Computational Linguistics.

\bibitem[{Zhang et~al.(2020)Zhang, Sun, Galley, Chen, Brockett, Gao, Gao, Liu, and Dolan}]{zhang-etal-2020-dialogpt}
Yizhe Zhang, Siqi Sun, Michel Galley, Yen-Chun Chen, Chris Brockett, Xiang Gao, Jianfeng Gao, Jingjing Liu, and Bill Dolan. 2020.
\newblock \href {https://doi.org/10.18653/v1/2020.acl-demos.30} {{DIALOGPT} : Large-scale generative pre-training for conversational response generation}.
\newblock In \emph{Proceedings of the 58th Annual Meeting of the Association for Computational Linguistics: System Demonstrations}, pages 270--278, Online. Association for Computational Linguistics.

\bibitem[{Zhao et~al.(2023)Zhao, Gella, Kim, Jin, Hazarika, Papangelis, Hedayatnia, Namazifar, Liu, and Hakkani-Tur}]{zhao-etal-2023-others}
Chao Zhao, Spandana Gella, Seokhwan Kim, Di~Jin, Devamanyu Hazarika, Alexandros Papangelis, Behnam Hedayatnia, Mahdi Namazifar, Yang Liu, and Dilek Hakkani-Tur. 2023.
\newblock \href {https://doi.org/10.18653/v1/2023.sigdial-1.28} {{``}what do others think?{''}: Task-oriented conversational modeling with subjective knowledge}.
\newblock In \emph{Proceedings of the 24th Annual Meeting of the Special Interest Group on Discourse and Dialogue}, pages 309--323, Prague, Czechia. Association for Computational Linguistics.

\bibitem[{Zhou et~al.(2018)Zhou, Prabhumoye, and Black}]{zhou-etal-2018-dataset}
Kangyan Zhou, Shrimai Prabhumoye, and Alan~W Black. 2018.
\newblock \href {https://doi.org/10.18653/v1/D18-1076} {A dataset for document grounded conversations}.
\newblock In \emph{Proceedings of the 2018 Conference on Empirical Methods in Natural Language Processing}, pages 708--713, Brussels, Belgium. Association for Computational Linguistics.

\end{thebibliography}
%\bibliography{custom}

\clearpage

\appendix

\section{Appendix}
\label{sec:appendix}

\subsection{Datasets}
\label{sec:app_datasets}
We briefly present the reasons for selecting the datasets.

\textbf{Open-Domain Dialogue (ODD)}
Differently from other datasets, DailyDialog dialogues only involve two participants~\citep{tiedemann-2009-opensubtitles, baumgartner-2020-pushshift-reddit}, are not audio transcriptions~\cite{godfrey-1992-switchboard}, have more than two exchanges between the participants~\citep{rashkin-etal-2019-towards}, and are not restricted by a persona (i.e. few sentences describing the user's interests)~\citep{zhang-etal-2018-personalizing, xu-etal-2022-beyond}.

\textbf{Knowledge-Grounded Dialogue (KGD)}
Wizard of Wikipedia provides a test set with an unseen set of documents~\citep{zhou-etal-2018-dataset, komeili-etal-2022-internet} and its knowledge has not changed over time (i.e. comparable with previous/future studies)~\citep{gopalakrishnan-2019-topical,hedayatnia-etal-2020-policy}.

\textbf{Task-Oriented Dialogue (TOD)}
A few other TOD datasets include unstructured knowledge access but consist only of a spoken test set~\citep{kim-2021-dstc10}, or provide no dialogue state annotation~\citep{feng-etal-2020-doc2dial}. 
The dataset proposed in the ninth Dialogue System Technology Challenge augmented MultiWOZ 2.1~\citep{eric-etal-2020-multiwoz} with knowledge access turns but removed the dialogue state annotation.
To always include the dialogue state in our analysis, we recovered the dialogue state annotation from the original MultiWOZ 2.1 dialogues, and we only considered the dialogues from this dataset.

\textbf{Question Answering (QA)}
We choose NarrativeQA because it has a publicly available test set (to evaluate the retriever) and answers are expressed as free-form text (to evaluate response generation)~\citep{rajpurkar-etal-2016-squad, rajpurkar-etal-2018-know, yang-etal-2018-hotpotqa, kwiatkowski-etal-2019-natural}.
Although the original task always provides the correct document, we also wanted to investigate the performance of the retriever when considering documents with an average length of 600 tokens.
Additionally, we avoided splitting documents into smaller chunks (e.g. passages or sentences) because this would have made the computation of the retriever performance more challenging.

\subsection{Implementation and resources}
\label{sec:app_impl_and_perf}
\textbf{Models and parameters}
We fine-tuned the models using LoRA (rank 32 and alpha 64) for a maximum of 10 epochs with an early stopping patience of 2.
We chose AdamW~\citep{loshchilov2017decoupled} as the optimizer and used a learning rate of $10^{-4}$ for \llama and $10^{-5}$ for \mistral (selected based on the performance on the development sets).
To obtain an encoding for both documents and queries, we used all-mpnet-base-v2\footnote{\url{https://www.sbert.net/docs/pretrained\_models.html}}. We have then stored the encoded documents in a FAISS vector store (used for retrieval).

\textbf{Input structure}
We separated the segments of the input vector with their name followed by a colon (i.e. "Dialogue state:", "Topic:", "Knowledge:", "Question:", "Answer:") similarly to previous work~\citep{izacard-grave-2021-leveraging, wang-2022-todnlg, chen-etal-2023-exploring-context, sun-etal-2023-generative}.
For TOD, we represented the dialogue state as a comma-separated list of domain slot value triplets~\citep{hosseini-asl-etal-2020-simpletod, wang-2022-todnlg}.

\textbf{Instructions}
Table \ref{tab:instructions} reports the instructions used for in-context learning experiments. For each dialogue type, we have experimented with three different instructions describing the task and the various input segments (e.g. dialogue history, topic, and knowledge). We have selected the best instruction based on the development set performance.

\begin{table*}[t!]
\small
    \centering
    \begin{tabularx}{\linewidth}{cX}
        \toprule
        \textbf{Dialogue Type} & \textbf{Instruction} \\
        \midrule
        \multirow{5}{*}{\textit{ODD}} & \texttt{""} \\
        \cmidrule{2-2}
        & \texttt{"This is a conversation between two people. Use the context to write an engaging reply for the other person."} \\
        \cmidrule{2-2}
        & \texttt{"Write a coherent continuation for the proposed conversation."} \\
        \midrule
        \multirow{6}{*}{\textit{KGD}} & \texttt{""} \\
        \cmidrule{2-2}
        & \texttt{"This is a conversation between two people about a Topic. Use the Dialogue and the additional Knowledge as context to write an engaging reply for the other person.",} \\
        \cmidrule{2-2}
        & \texttt{"Write a coherent continuation for the proposed conversation based on the additional Knowledge."} \\
        \midrule
        \multirow{5}{*}{\textit{TOD}} & \texttt{""} \\
        \cmidrule{2-2}
        & \texttt{"In the following conversation a user wants to achieve some goal and needs help from an assistant. Continue the conversation with the response of the assistant."} \\
        \cmidrule{2-2}
        & \texttt{"Write a coherent continuation for the proposed conversation."} \\
        \midrule
        \multirow{5}{*}{\textit{QA}} & \texttt{""} \\
        \cmidrule{2-2}
        & \texttt{"You are presented with a user's Question about a movie or book. Answer to the user's Question using the information provided in the Context."} \\
        \cmidrule{2-2}
        & \texttt{"Answer to the user's question using the provided information (if available)."} \\
        \bottomrule
    \end{tabularx}
    \caption{
        Instructions used to adapt the model to a specific dialogue type with in-context learning. We defined three instructions for each dialogue type, describing the task and the various input segments (e.g. dialogue history, topic, dialogue state, and knowledge). We selected the best instruction based on the development set performance.
    }
    \label{tab:instructions}
\end{table*}

\textbf{Generation}
We sampled 10\% of the data (in a stratified fashion, based on the length of the responses) from the development set of each dialogue type.
For each model, we used grid search to find, for the sampled data, the combination of parameters (top-p, top-k, and temperature) leading to the highest BLEU-4.
The best combination of parameters was used to generate the responses for the test set.

\textbf{GPU Requirements}
Most computations were performed on a single NVIDIA A100 GPU with 80GB, requiring less than 50 hours to execute.
In a few cases, we had to use two (i.e. fine-tuning the models for QA using more than one document) or three (i.e. integrated gradients) A100 with 80GB each.

\subsection{Additional Automatic Evaluation}
\label{sec:add_autom_eval}

To automatically evaluate the quality of the generated text, we have considered BLEU-4 \cite{papineni-etal-2002-bleu}, F1 (i.e. unigram overlap), and ROUGE-L \cite{lin-2004-rouge}. Furthermore, we have used KF1 \cite{shuster-etal-2021-retrieval-augmentation} to measure the overlap between the prediction and the knowledge selected by the annotators.
For reproducibility purposes, we have computed ROUGE-L using the official implementation\footnote{\url{https://github.com/google-research/google-research/tree/master/rouge}} and all the remaining metrics using ParlAI\footnote{\url{https://parl.ai}}. No pre-processing was performed on the model-generated answers.

Table \ref{tab:auto_eval} reports the performance for each dialogue type. As mentioned in Section \ref{sec:auto-eval}, the best performance is obtained by fine-tuned models. 
Following, we analyze the results for each dialogue type.

\textbf{Open-Domain Dialogue (ODD)}
Although fine-tuning achieves a higher BLEU-4, the results show that both techniques produce very different responses with respect to the ground truth.

\begin{table*}[ht!]
\small
    \centering
    \begin{adjustbox}{width=1.03\linewidth,center=\linewidth}
        \begin{tabularx}{1.03\linewidth}{lllccccccc}
            \toprule
            \multirow{2}{*}{\textbf{Model}} & \multirow{2}{*}{\textbf{Technique}} & \multirow{2}{*}{\textbf{\makecell{External\\Knowledge}}} & \multicolumn{2}{c}{\textbf{BLEU-4}} & \multicolumn{3}{c}{\textbf{KF1}} & \textbf{F1} & \textbf{ROUGE-L} \\
            \cmidrule(rrr){4-5} \cmidrule(rrr){6-8} \cmidrule(r){9-9} \cmidrule(r){10-10}
            & & & \texttt{ODD} & \texttt{TOD} & \texttt{KGD} & \texttt{TOD}  & \texttt{QA} & \texttt{KGD} & \texttt{QA} \\
            \midrule
            \multirow{6}{*}{\textbf{\llama}} & \multirow{3}{*}{\textit{In-Context Learning}} & \texttt{No Know.} & 0.2 & 0.85 & 11.61 & 13.66 & 5.26 & 12.68 & 5.59 \\
            & & \texttt{Retrieved Know.} & & 0.83 & 13.51 & 12.10 & 5.65 & 12.91 & 14.86 \\
            & & \texttt{Gold Know.} & & 1.07 & 25.87 & 21.03 & \textbf{6.72} & 16.59 & 23.22 \\
            \cmidrule{2-10}
             & \multirow{3}{*}{\textit{Fine-Tuning}} &\texttt{No Know.} & \textbf{0.3} & \textbf{6.72} & 17.43 & 34.04 & 0.74 & 18.46 & 17.25 \\
            & & \texttt{Retrieved Know.} & & 4.33 & 25.10 & 26.85 & 1.15 & 20.70 & 46.21 \\
            & & \texttt{Gold Know.} & & 5.39 & \textbf{76.23} & \textbf{42.69} & 1.44 & \textbf{38.41} & \textbf{73.38} \\
            \midrule
            \multirow{6}{*}{\textbf{\mistral}} & \multirow{3}{*}{\textit{In-Context Learning}} & \texttt{No Know.} & 0.2 & 1.33 & 10.96 & 13.01 & 4.84 & 11.04 & 6.94 \\
            & & \texttt{Retrieved Know.} & & 1.06 & 13.83 & 12.53 & 6.09 & 12.22 & 10.26 \\
            & & \texttt{Gold Know.} & & 1.33 & 25.95 & 28.74 & \textbf{7.07} & 15.88 & 21.74 \\
            \cmidrule{2-10}
             & \multirow{3}{*}{\textit{Fine-Tuning}} &\texttt{No Know.} & \textbf{0.9} & \textbf{4.09} & 15.47 & 29.27 & 0.67 & 18.63 & 12.73 \\
            & & \texttt{Retrieved Know.} & & 3.85 & 21.63 & 30.44 & 1.18 & 20.49 & 45.40 \\
            & & \texttt{Gold Know.} & & 3.94 & 68.36 & \textbf{43.04} & 1.46 & \textbf{38.21} & \textbf{70.54} \\   
            \midrule
            \textbf{Ground Truth} & & & 100 & 100 & 37.79 & 38.48 & 1.52 &  100 & 100 \\
            \bottomrule
        \end{tabularx}
    \end{adjustbox}
    \caption{ \textbf{Automatic Evaluation} BLEU-4, KF1, F1 and ROUGE-L for In-Context Learning and Fine-Tuning with \texttt{Retrieved} (top-3) and \texttt{Gold} (ground-truth) knowledge, on \llama and \mistralp, in different dialogue types: Open-Domain Dialogues (ODDs), Knowledge Grounded Dialogues (KGDs), Task-Oriented Dialogues (TODs), and Question Answering (QA). }
    \label{tab:auto_eval}
\end{table*}

\begin{table*}[ht!]
\small
    \centering
    \begin{tabularx}{0.8\linewidth}{lllcccc}
        \toprule
        \multirow{2}{*}{\textbf{Model}} & \multirow{2}{*}{\textbf{Technique}} & \multirow{2}{*}{\textbf{\makecell{External\\Knowledge}}} & \multicolumn{2}{c}{\textbf{BLEU-4}} & \multicolumn{2}{c}{\textbf{KF1}} \\
        \cmidrule(rr){4-5} \cmidrule(rr){6-7}
        & & & \texttt{TOD} & \texttt{TOD}\textsuperscript{\textdagger} & \texttt{TOD} & \texttt{TOD}\textsuperscript{\textdagger} \\
        \midrule
        \multirow{6}{*}{\textbf{\llama}} & \multirow{3}{*}{\textit{In-Context Learning}} & \texttt{No Know.} & 0.85 & 0.60 & 13.66 & 12.39  \\
        & & \texttt{Retrieved Know.} & 0.83 & 0.44 & 12.10 & 10.44  \\
        & & \texttt{Gold Know.} & 1.07 & 2.67 & 25.87 & 23.77  \\
        \cmidrule{2-7}
         & \multirow{3}{*}{\textit{Fine-Tuning}} &\texttt{No Know.} & \textbf{6.72} & 4.33 & 34.04 & 25.73 \\
        & & \texttt{Retrieved Know.} & 4.33 & 3.15 & 26.85 & 22.92  \\
        & & \texttt{Gold Know.} & 5.39 & \textbf{8.50} & \textbf{42.69} & \textbf{45.49}  \\
        \midrule
        \multirow{6}{*}{\textbf{\mistral}} & \multirow{3}{*}{\textit{In-Context Learning}} & \texttt{No Know.} & 1.33 & 1.12 & 13.01 & 11.91  \\
        & & \texttt{Retrieved Know.} & 1.06 & 1.02 & 12.53 & 10.36  \\
        & & \texttt{Gold Know.} & 1.33 & 3.70 & 28.74 & 28.79  \\
        \cmidrule{2-7}
         & \multirow{3}{*}{\textit{Fine-Tuning}} &\texttt{No Know.} & \textbf{4.09} & 5.83 & 29.27 & 25.47  \\
        & & \texttt{Retrieved Know.} & 3.85 & 4.76 & 30.44 & 25.61  \\
        & & \texttt{Gold Know.} & 3.94 & \textbf{10.63} & \textbf{43.04} & \textbf{49.40}  \\        
        \midrule
        \textbf{Ground Truth} & & & 100 & 100 & 38.48 & 39.91 \\
        \bottomrule
    \end{tabularx}
    \caption{
        \textbf{Automatic Evaluation}
        BLEU-4 and KF1 for In-Context Learning and Fine-Tuning with \texttt{Retrieved} (top-3) and \texttt{Gold} (ground-truth) knowledge, on \llama and \mistralp, in Task-Oriented Dialogues (TODs). 
        \textsuperscript{\textdagger} indicates that only test turns with unseen knowledge were included.
    }
    \label{tab:tod_auto_eval}
\end{table*}

\textbf{Knowledge-Grounded Dialogue (KGD)}
We report the performance of the models on the unseen test set (i.e. the knowledge base contains documents that are only present in the test set).
The results show that models adapted using fine-tuning obtain a higher F1 than in-context learning.
Furthermore, the best models tend to copy more from the gold knowledge compared to the annotators (as shown in the ground truth).

\textbf{Task-Oriented Dialogue (TOD)}
Differently from the other types, \llama and \mistral have obtained the best performance in terms of BLEU-4 when fine-tuned with no additional knowledge.
Further investigation suggests this happens because of the high overlap between the knowledge used for training and testing (82\%).
We report the performance on the documents only available in the test phase in Table \ref{tab:tod_auto_eval} (\texttt{TOD\textsuperscript{\textdagger}}). In this scenario, gold knowledge does indeed increase the performance of the models.

\textbf{Question Answering (QA)}
Although fine-tuned models achieve the highest ROUGE-L, in-context learning models tend to provide longer and possibly more detailed responses, as reported in terms of KF1.
Because ground truths are particularly short (4.26 tokens on average), models that generated longer responses (especially models adapted with in-context learning) were awarded a lower ROUGE-L.

\subsubsection{Retriever Accuracy}
\begin{figure}[ht!]
    \centering
    \includegraphics[width=\columnwidth]{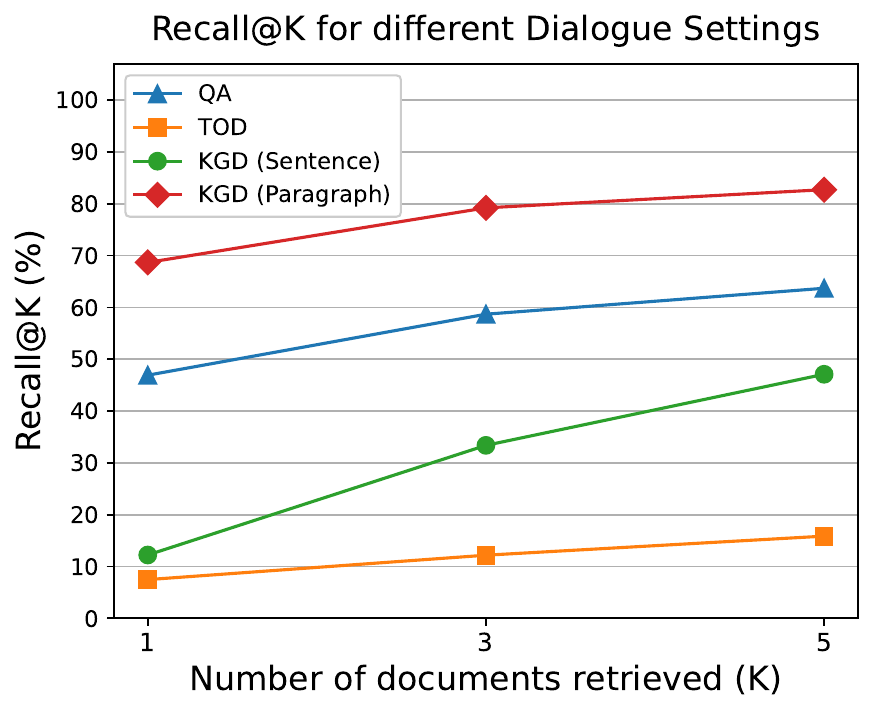}
    \caption{
        Performance of the off-the-shelf retriever for each dialogue type. 
        The retriever achieves the lowest Recall@K on TOD because of the larger knowledge base size (2900 documents). 
        However, the retriever achieves a higher Recall@K for QA, even though its knowledge base is bigger than the one for KGD (355 vs. 61 $\pm$ 21).
        Further studies indicate that, despite the model is not capable to retrieve the exact sentence of the annotator (KGD Sentence), the retriever selects a sentence belonging to the same paragraph more than 69\% of the time (KGD Paragraph).
    }
    \label{fig:recall-at-k}
\end{figure}

We study the performance of the retriever for each dialogue type and report Recall@K in Figure \ref{fig:recall-at-k}.
Because of the size of the knowledge base (Table \ref{tab:datasets}), the retriever achieves the lowest performance on TOD.
However, although the knowledge base for QA is bigger than for KGD, the retriever achieves a higher recall for QA.
Further study suggest that, although the retriever selects the gold sentence in only a few cases, the model retrieves a sentence from the same paragraph more than 69\% of the time.

\subsection{Human Evaluation}
\label{sec:app_hum_eval}
Table \ref{tab:human_eval-appendix} reports the results for the "Correctness" dimension of Human Evaluations. Except for ODD, fine-tuning tends to improve correctness.

Table \ref{tab:hum_eval_validity_question} presents the question and the answer options for the proposed "Validity" dimension used in QA.

\begin{table*}[ht!]
\small
    \centering
    \begin{tabularx}{0.8\linewidth}{lXXcccc}
        \toprule
        \multirow{2}{*}{\textbf{Model}} & \multirow{2}{*}{\textbf{Technique}} & \multirow{2}{*}{\textbf{\makecell{External\\Knowledge}}} & \multicolumn{4}{c}{\textbf{Correctness}}  \\
        \cmidrule(rrrr){4-7} 
        & & & \texttt{ODD} & \texttt{KGD} & \texttt{TOD} & \texttt{QA} \\
        \midrule
        \multirow{6}{*}{\textbf{\llama}} & \multirow{3}{*}{\textit{In-Context Learning}} & \texttt{No Know.} & \textbf{95} & 80 & \textbf{95} & 75 \\
        & & \texttt{Retrieved Know.} & & 80 & 60 & 60  \\
        & & \texttt{Gold Know.} & & 80 & 70 & 80 \\
        \cmidrule{2-7}
        & \multirow{3}{*}{\textit{Fine-Tuning}} & \texttt{No Know.} & 65 & \textbf{90} & 70 & 75 \\
        & & \texttt{Retrieved Know.} & & \textbf{90} & 90 & 55 \\
        & & \texttt{Gold Know.} & & 85 & 85 & \textbf{85} \\
        \midrule
        \multirow{6}{*}{\textbf{\mistral}} & \multirow{3}{*}{\textit{In-Context Learning}} & \texttt{No Know.} & \textbf{95} & 70 & 75 & 60 \\
        & & \texttt{Retrieved Know.} & & 55 & 70 & 50 \\
        & & \texttt{Gold Know.} & & \textbf{85} & 60 & 80 \\
        \cmidrule{2-7}
        & \multirow{3}{*}{\textit{Fine-Tuning}} & \texttt{No Know.} & 65 & \textbf{85} & 80 & 50 \\
        & & \texttt{Retrieved Know.} & & 75 & \textbf{100} & 45 \\
        & & \texttt{Gold Know.} & & 70 & 80 & \textbf{85} \\
        \midrule
        \textbf{Ground-Truth} & & & 95 & 70 & 85 & 80 \\
        \bottomrule
        \end{tabularx}
        \caption{
            \textbf{Human Evaluation}
            Percentage of Correct (ODD, KGD, TOD, QA) responses for In-Context Learning and Fine-Tuning with \texttt{Retrieved} (top-3) and \texttt{Gold} (ground-truth) knowledge, on \llama and \mistralp, for different dialogue types: Open-Domain Dialogues (ODDs), Knowledge Grounded Dialogues (KGDs), Task-Oriented Dialogues (TODs), and Question Answering (QA). 
        }
    \label{tab:human_eval-appendix}
\end{table*}

\begin{table*}[t!]
\small
    \centering
    \begin{tabularx}{\linewidth}{cccX}
        \toprule
        \textbf{Dimension} & \textbf{Question} & \textbf{Answer Option} & \textbf{Option Definition} \\
        \midrule
        \multirow{7}{*}{\textbf{Validity}} & \multirow{7}{*}{\textit{\makecell{Is the response \\ candidate valid?}}} & \multirow{2}{*}{\texttt{Valid}} & The response candidate includes the right information from the context to adequately answer the proposed question. \\
        \cmidrule{3-4}
        & & \multirow{2}{*}{\texttt{Not Valid}} & The response candidate does not include the right information from the context to adequately  answer the proposed question. \\
        \cmidrule{3-4}
        & & \multirow{2}{*}{\texttt{I don't know}} & The response candidate includes some information that is adequate to answer the proposed question, but some that is not. \\
        \bottomrule
    \end{tabularx}
    \caption{
        Question and answer options presented to the annotators for the proposed Validity dimension.
    }
    \label{tab:hum_eval_validity_question}
\end{table*}

\end{document}